\documentclass[nonatbib,final,5p,times,twocolumn]{elsarticle}

\let\OLDthebibliography\thebibliography
\renewcommand\thebibliography[1]{
  \OLDthebibliography{#1}
    \setlength{\itemsep}{0pt plus .5pt}
  \footnotesize \vskip 0.3\baselineskip plus 0.1\baselineskip minus 0.1\baselineskip%
}

\usepackage{graphics}
\usepackage{epsfig}
\usepackage{times}
\usepackage{amsmath}

\usepackage{cite}

\usepackage{subcaption}
\usepackage{xargs}
\usepackage[normalem]{ulem}
\usepackage[ruled,vlined,linesnumbered]{algorithm2e}
\usepackage{textcomp}
\usepackage{multirow}
\usepackage{lipsum}
\usepackage{arydshln}
\usepackage{bm}
\usepackage{float}
\usepackage{comment}
\usepackage{todonotes}

\usepackage{amssymb}

\newcommand{\xchange}[2]{#2}

\DeclareMathOperator*{\argmax}{arg\,max}

\DeclareMathOperator{\E}{\mathbb{E}}

\DeclareMathOperator{\Cov}{\widehat{Cov}}
\newcommand{\diag}{\mathop{\mathrm{diag}}}

\journal{Robotics and Autonomous Systems}
\begin{document}
\begin{frontmatter}



\title{DROPO: Sim-to-Real Transfer with Offline Domain Randomization}


\author[inst1,inst2]{Gabriele Tiboni}

\affiliation[inst1]{organization={Dept. of Control and Computer Engineering, Politecnico di Torino},
            addressline={Francesco Ferrucci Street, 112}, 
            city={Torino},
            postcode={10141}, 
            country={Italy}}

\author[inst2]{Karol Arndt\corref{cor1}}
\ead{karol.arndt@aalto.fi}
\author[inst2]{Ville Kyrki}
\affiliation[inst2]{organization={Intelligent Robotics Group, Dept. of Electrical Engineering and Automation, Aalto University},
            addressline={Maarintie 8}, 
            city={Espoo},
            postcode={02150}, 
            country={Finland}}
\cortext[cor1]{Corresponding author.}

\begin{abstract}
In recent years, domain randomization over dynamics parameters has gained a lot of traction as a method for sim-to-real transfer of reinforcement learning policies in robotic manipulation; however, finding optimal randomization distributions can be difficult.
In this paper, we introduce DROPO, a novel method for estimating domain randomization distributions for safe sim-to-real transfer.
Unlike prior work, DROPO only requires a limited, precollected offline dataset of trajectories, and explicitly models parameter uncertainty to match real data using a likelihood-based approach.
We demonstrate that DROPO is capable of recovering dynamic parameter distributions in simulation and finding a distribution capable of compensating for an unmodeled phenomenon.
We also evaluate the method in two zero-shot sim-to-real transfer scenarios, showing successful domain transfer and improved performance over prior methods.
\end{abstract}


\end{frontmatter}


\section{Introduction}
\label{sec:introduction}
Over the past decade, there have been significant research advances in reinforcement learning (RL), resulting in a variety of success stories, from playing Atari games~\cite{mnih2013atari} to defeating the world champion in Go~\cite{Silver16alphago}.
These successes have sparked a lot of interest in RL among roboticists;
however, due to low sample efficiency of most algorithms, the challenge of accessing significant data collections at training time, as well as the need for random exploration resulting in the risk of hardware damage, applications of RL in robotics have been lagging behind.

A promising approach to solve the problem of limited data availability is to use simulated environments for training the RL agents, and to later deploy them on real-world physical systems~\cite{zhao2020simtoreal,tan2018sim,sadeghi2017cad2rl}.
However, while this solves the problem of fast and safe data collection, inaccuracies in simulation caused by, e.g., uncertainty over physical parameters and unmodeled phenomena, may result in policies that do not transfer well to the real system. Such discrepancy is commonly known as the \textit{reality gap}.

One can aim for more generalizable RL agents by performing the training over multiple variants of the environment, varying in appearance~\cite{sadeghi2017cad2rl,hamalainen2019affordance,Tobin2017domain,sadeghi17sim2real}, dynamics parameters~\cite{peng2018sim, arndt2019meta, tan2018sim, pivotingkragic, Nagabandi2019learning, transferMurtaza}, or both~\cite{andrychowicz19learning,du_auto-tuned_2021}.
This technique, known as \textit{domain randomization} (DR), has proven to be successful in a range of robotics setups.
However, finding well-performing domain randomization distributions over physical parameters can be difficult~\cite{valassakis2020crossing}---the distributions need to be wide enough to encompass the plausible true parameter values in the real world, or make up for phenomena which are specific to the real system but are not modelled in the simulator~\cite{arndt2020fewshot}. At the same time, the DR distribution needs to allow for convergence while training a single policy that performs well across all the randomized environments, especially in the zero-shot transfer scenario.
Therefore, there has been an increasing research interest in developing approaches to automate DR~\cite{rajeswaran_epopt_2017, bayessimramos,tsai2021droid,chebotar19closing,mehta20active}.

\begin{figure}
    \centering
    \includegraphics[width=0.8\linewidth]{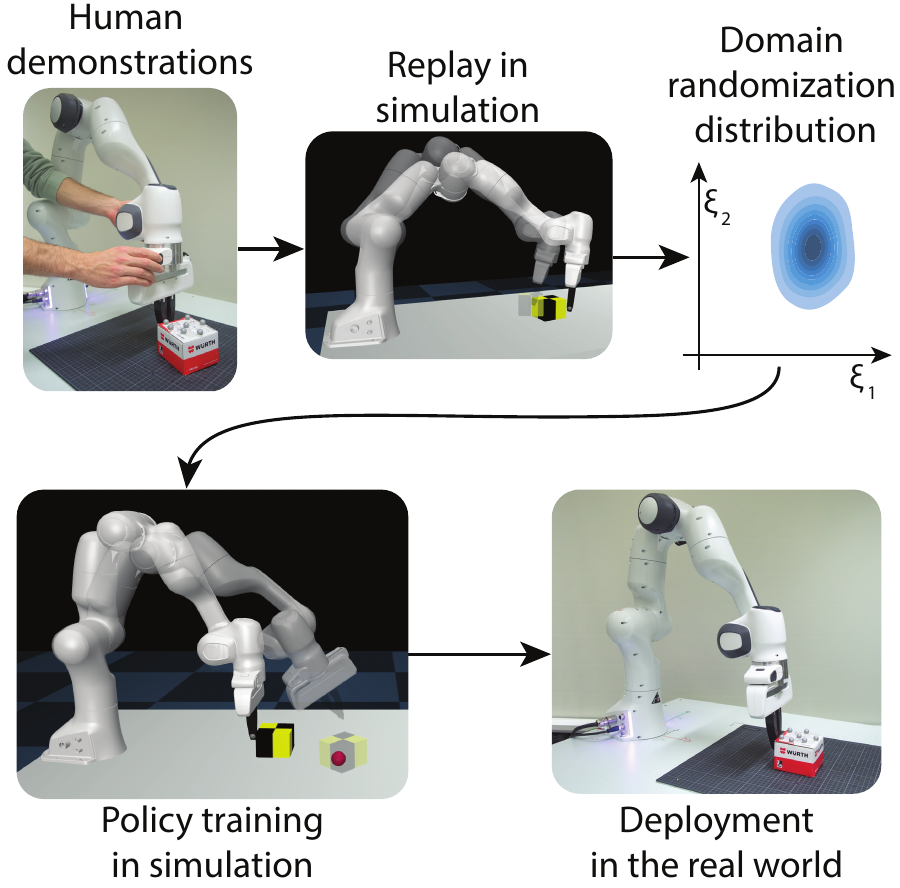}
    \caption{DROPO uses off-policy data or human demonstrations to learn a domain randomization distribution, which is later used to train a policy that can be directly transferred to the real world.}
    \label{fig:dropo_main}
\end{figure}

In this paper, we introduce Domain Randomization Off-Policy Optimization (DROPO), a novel method for automatically inferring domain randomization distributions.
DROPO fits a dynamics parameter distribution to an offline dataset using a likelihood-based metric, which explicitly maximizes the probability of observing real-world data in simulation. 
The simulator can thus be thought of as a stochastic forward model whose randomness arises from variability in the physical parameters of the scene.
Our approach, outlined in Figure~\ref{fig:dropo_main}, 
departs the trend of online-adaptive DR---where data from the target domain is collected on-policy during the parameter optimization process, requiring iterative access to the real-world setup~\cite{Muratore2021DataEfficientDR,chebotar19closing,rajeswaran_epopt_2017,du_auto-tuned_2021}---and follows the recently introduced offline setting~\cite{tsai2021droid}, while using a probabilistic metric to encourage variance in the converged parameter distributions.
This allows DROPO to optimize DR distributions using only precollected data (e.g. human demonstrations) without suffering from the distributions collapsing to point estimates due to L2-based loss functions.
In contrast to previous claims~\cite{bayessimramos}, our findings support that a likelihood-based method can be effortlessly implemented with a non-differentiable black-box simulator and gradient-free optimization techniques.

The contributions of this work are:
(1) introducing DROPO, a novel method for offline optimization of domain randomization distributions given safely collected data,
(2) an experimental evaluation of DROPO to show convergence to ground truth dynamics distributions,
(3) demonstrating that domain randomization distributions obtained with DROPO can be used to successfully train control policies in the presence of unmodeled phenomena,
(4) showing that RL policies can be directly transferred to the real world when trained in simulation on DROPO's dynamics distributions.
The source code for running DROPO with arbitrary offline datasets is publicly available\footnote{https://github.com/gabrieletiboni/dropo}.

\section{Related work}
\label{sec:relatedwork}
Domain randomization as a method for training generalizable reinforcement learning policies has been widely studied over the past few years, both in the context of computer vision~\cite{Tobin2017domain, sadeghi17sim2real, sadeghi2017cad2rl, hamalainen2019affordance}, dynamics models~\cite{peng2018sim, arndt2019meta, tan2018sim, pivotingkragic, Nagabandi2019learning, transferMurtaza} and both at the same time~\cite{andrychowicz19learning,du_auto-tuned_2021}.
A major challenge, however, is that domain randomization requires the randomization distributions for each parameter to be specified when the model is trained.
DR essentially trades optimality for robustness: using too wide ranges may have negative impact on the training, making it difficult for the agent to learn a single policy that generalizes to such a wide range of dynamics, while using too narrow ranges would bias the model towards a small set of dynamics values and thus possibly hinder generalization to the real world, as in the case of system identification of point-estimate dynamics~\cite{kolev_physically_2015,allevato_iterative_2020}.

To this end, multiple methods for automatically tuning the randomization ranges were proposed,
such as minimizing discrepancy between simulated and real trajectories~\cite{chebotar19closing} or directly maximizing the expected real-world return~\cite{Muratore2021DataEfficientDR};
however, as these methods explicitly ask for real-world feedback during the process, they require access to and careful supervision of the physical setup in order to run the policy trained on the intermediate DR distributions.

To relax these requirements, 
DROID~\cite{tsai2021droid} proposed a setting where the DR distribution is optimized using a fixed, offline dataset pre-collected in the target domain, aiming to improve data-efficiency and reduce safety risks coming from rolling out policies trained in unconverged dynamics in contact-rich tasks.
This framework reduced the real-world interaction to a single data collection step performed before parameter optimization; additionally, this setting does not make any assumptions about the data collection procedure, allowing it to be collected with another policy, or through human demonstrations.
Our method follows the same problem setting, but differs from DROID in that it uses a likelihood-based objective function and optimizes the distribution variance in addition to the mean.
We provide a thorough comparison between our method and DROID in Section~\ref{sec:experiments}.

A variation of this offline setting, where the real-world data collection policy is assumed to be available, was addressed in BayesSim~\cite{bayessimramos}. Under this assumption, BayesSim trains a neural network to predict dynamics parameter posteriors using trajectories collected in simulation by the given policy under a variety of dynamic conditions.
Trajectories collected by the same policy on the real world setup are then passed to the trained network to predict the dynamics parameter distributions in the real world.
Similar to our work and DROID, BayesSim does not require access to the real hardware during optimization; however, it still assumes the data collection policy to be available during optimization. 
In contrast, we propose a maximum likelihood-based framework which does not require the data collection policy to be known, making it more suitable for use with human demonstrations.
We provide a detailed comparison between our method and BayesSim with human demonstrations in Section~\ref{sec:experiments}.

Other works followed a different approach for applying domain randomization,
by keeping a fixed range of the physical parameters and, instead, relying on intelligent sampling techniques to improve generalization: \cite{mehta20active} guides training on increasingly harder environment variations, while~\cite{murattore19assessing} increases the number of sampled simulated environments until satisfactory transfer behavior is reached.

Finally, approaches based around differentiable physics simulators were also devised~\cite{deavila2018differentiable}.
While DROPO could in principle be used with a differentiable simulator, allowing for the use of a gradient-based optimizer, we do not make any assumptions about the differentiability of the simulator in this work.

\section{Method}
\label{sec:method}
\subsection{Background and problem formulation}
A Markov decision process (MDP) $\mathcal{M}$ is described by its state space $\mathcal{S}$, action space $\mathcal{A}$, initial state distribution $p(s_0)$, state transition distribution $p(s_{t+1}|s_{t}, a_{t})$ and the reward distribution $p(r_{t}|s_{t}, a_{t}, s_{t+1})$.
Each \textit{episode} begins with an \textit{agent} starting in state $s_0 \sim p(s_0)$.
At each step, the agent selects an action $a$ following a policy $\pi(a|s)$.
The problem addressed in reinforcement learning is then to find an optimal policy $\pi^*(a|s)$, such that the expected cumulative reward (the \textit{return}) is maximized.

In the domain transfer scenario, we can view the real world as an MDP with state transition probabilities $p_{real}(s_{t+1}|s_{t}, a_{t})$, with the simulation additionally parametrized by the dynamics parameter vector $\xi$: $p_{sim}(s^{sim}_{t+1}|s_{t}, a_{t}, \xi)$.
We assume that both MDPs share the state and action spaces, the initial state distribution, and the reward function. 

Under this formulation, the problem we address in this work can be stated as follows: given a dataset of real-world state-action trajectories $\mathcal{D} = \{ s_0, a_0, s_1, \ldots, s_T \}$, find the distribution $p^*(\xi)$ such that the likelihood of transitioning to the real state $s_{t+1}$ when taking action $a_t$ while in $s_{t}$ is maximized over the dataset:

 \begin{equation}
     p^*(\xi) = \argmax_{p(\xi)} \mathop{\E}_{\substack{s_{t}, a_t, s_{t+1} \sim \mathcal{D} \\ \xi \sim p(\xi)}} p_{sim}(s_{t+1} | s_{t}, a_t, \xi).
 \end{equation}

\subsection{Method overview}
\begin{figure}
    \centering
    \includegraphics[width=0.95\linewidth]{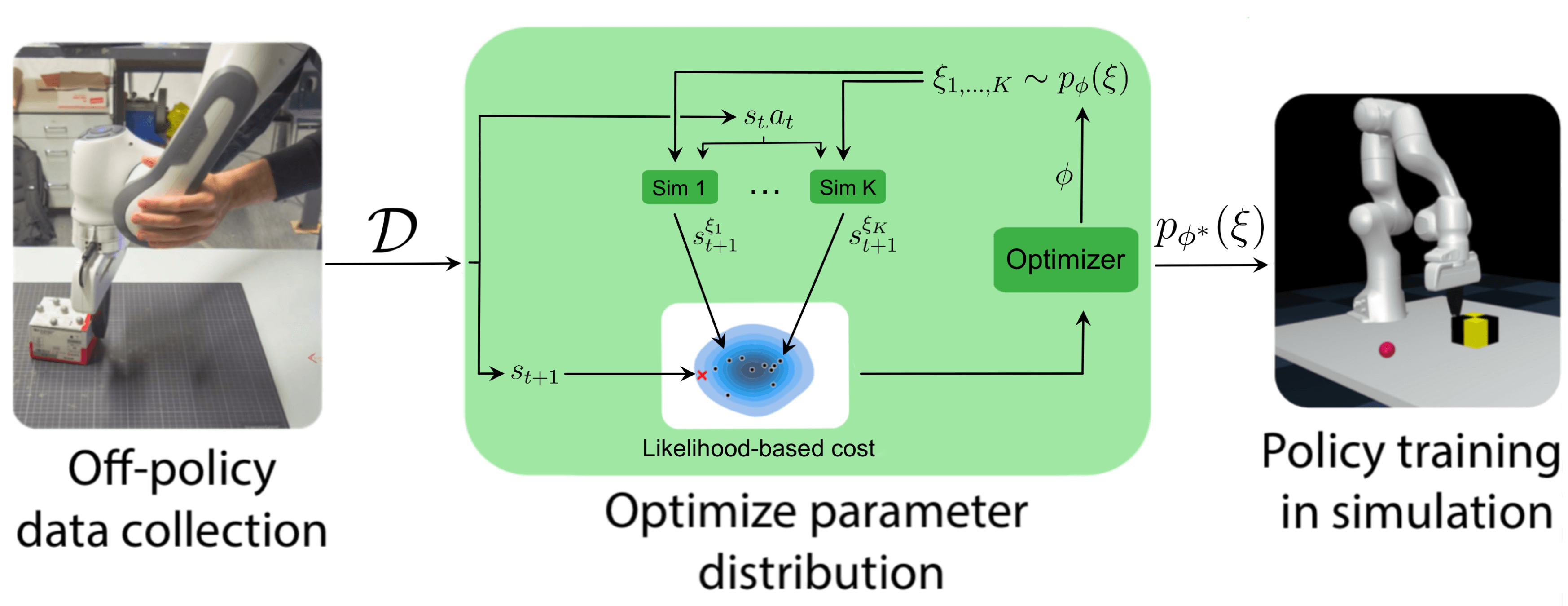}
    \caption{\xchange{(Figure modified)}{}Overview of the main stages of DROPO. The dynamics parameter distribution is optimized to maximize the probability of observing the collected dataset in simulation.}
    \label{fig:overview}
\end{figure}

DROPO consists of three main steps: dataset collection on the real hardware, dynamics distribution fitting, and policy training.
An overview of the method is presented in Figure~\ref{fig:overview} with a step-by-step walkthrough of dynamics distribution fitting presented in Algorithm~\ref{alg:dropo}.

\begin{algorithm}
\SetAlgoLined  
\KwResult{Parameters $\phi^*$ of $p_{\phi^*}(\xi)$}
 Initialize $\phi$ to $(\mu_{init},\Sigma_{init})$  \; \label{as:init_theta}
 Initialize empty dataset $\mathcal{D}$ \; \label{as:init_d}
 Fill $\mathcal{D}$ with trajectories from the target domain \; \label{as:collect}
 \While{not converged}
 { \label{as:sample_xi} 
  Sample a set $\Xi$ of $K$ random dynamics $\xi \sim p_\phi(\xi)$ \; \label{as:sample}
  \ForAll{$s_{t}, a_{t..t+\lambda-1}, s_{t+\lambda} \in \mathcal{D}$}
  { \label{s:for_state}
    \ForAll{$\xi_k \in \Xi$}
    { \label{as:for_xi}
         Set simulator parameters to $\xi_k$ \;  \label{as:set_params}
         Set the simulator state to $s_{t}$ \;  \label{as:set_state}
         Execute actions $a_{t..t+\lambda-1}$ \;  \label{as:exec_actions}
         Observe $s^{\xi_k}_{t+\lambda} \sim p_{sim}(s^{sim}_{t+\lambda} | s_{t}, a_{t..t+\lambda-1}, \xi_k)$ \;  \label{as:observe_next_state}
     } \label{as:end_for_xi}
     Compute the mean $\bar{s}^{\phi}_{t+\lambda}$ and covariance $\Sigma^{\phi}_{t+\lambda}$ (Eq. \ref{eq:meanvar})\;  \label{as:fit_gaussian}
     Evaluate the log-likelihood $\mathcal{L}_{t}$ of $s_{t+\lambda}$ under $\mathcal{N}(\bar{s}^{\phi}_{t+\lambda}, \Sigma^{\phi}_{t+\lambda})$ (Eq. \ref{eq:loss})\;  \label{as:get_log_likelihood}
  }
  Compute the total log-likelihood $\mathcal{L} = \sum_{t} \mathcal{L}_t$\;  \label{as:get_total_log_likelihood}
  Update $\phi$ towards maximizing $\mathcal{L}$ \;  \label{as:update_phi}
 }
 Train a policy with DR using the converged $p_{\phi^*}(\xi)$ \;  \label{as:train_policy}
 \caption{DROPO}
 \label{alg:dropo}
\end{algorithm}

The data collection step is where the dataset $\mathcal{D}$, used for parameter optimization, is collected on the physical hardware.
This can be done either by running any previously trained policy (e.g., a policy for another task) or by manually guiding the robot through kinesthetic teaching. 
Since DROPO uses offline, off-policy data, there is no need to collect additional data at later stages.
This is also the only step where the physical hardware is necessary---all later steps of the method are performed in simulation, up until the final deployment.

The core part of DROPO is the second step, where the domain randomization distribution $p_\phi(\xi)$---parameterized by $\phi$---is optimized to maximize the likelihood of real-world data.
In particular, the likelihood is computed w.r.t. the \textit{next-state distribution} under $\phi$, which can be expressed as the marginal probability density
    $p_{sim}(s^{sim}_{t+1} | s_t, a_t, \phi) = \int p_{sim} (s^{sim}_{t+1} | s_t, a_t, \xi) p_\phi(\xi) d\xi$.
To approximate this integral, the state-action pair $(s_t, a_t)$ in question is executed multiple times through simulators with different dynamics parameters $\xi_k$ sampled from $p_\phi(\xi)$ (steps \ref{as:for_xi}--\ref{as:end_for_xi} in Algorithm~\ref{alg:dropo}).
By doing so, the resulting next-state distribution $p_{sim}(s^{sim}_{t+1} | s_{t}, a_{t}, \phi)$ may be estimated, which captures uncertainty induced by both the simulator dynamics and the given DR distribution.
The parameters $\phi$ of $p_\phi(\xi)$ are then adjusted to maximize the likelihood of the real $s_{t+1}$ under $p_{sim}(s^{sim}_{t+1} | s_{t}, a_{t}, \phi)$.

We can further generalize the algorithm by allowing to step the simulator by more than a single step during the inference process.
This mechanism allows to pick a desired evaluation frequency---regardless of the sampling rate of collected data---and to increase the signal-to-noise ratio in the difference between successive states.
This results in an additional hyperparameter $\lambda \in \mathbb{N}^+$, which controls how many successive actions are executed in the environment before evaluating the log-likelihood of the real $s_{t+\lambda}$ under $p_{sim}(s^{sim}_{t+\lambda} | s_{t}, a_{t..t+\lambda-1}, \phi)$. 
This generalization is reflected in step~\ref{as:exec_actions} of Algorithm~\ref{alg:dropo}.

Once the parameter optimization process has converged, the resulting distribution $p_{\phi^*}(\xi)$ is used to train a policy in the third step of DROPO.
This policy can then be deployed on the setup that the data was collected from in the first step.

\subsection{Data collection}
The first step of DROPO is to collect the state-action trajectory dataset $\mathcal{D}$, which is going to be used for domain randomization parameter optimization.
This dataset can either be collected from human demonstrations, or by running a previously trained, safe policy in the environment.
Regardless of the method used, the data should be collected in such a way that allows for identification of the relevant dynamics parameters;
for example, it will not be possible for DROPO to infer the friction coefficient of an object that does not move during demonstrations.

In case demonstrations are provided, the true actions are not directly known and need to be inferred based on the measured state sequence using an inverse dynamics model.
While learning inverse dynamics may be a challenging problem in itself, in certain environments---such as position- or velocity-controlled robot arms---this inference can be simplified under the assumptions that the robot's on-board controllers are perfect.
While this is never the case in real-world systems, we found that DROPO was able to handle this discrepancy.

Once collected, the dataset is preprocessed in order for the states to be directly replicable in simulation.
First, data from different sensor modalities needs to be synchronized;
this is because data from position measuring devices, such as motion capture systems, is likely to be delayed w.r.t. robot joint position measurements.
Additionally, it is necessary to ensure that the data is sampled with the same frequency as the simulated environment timesteps.
To achieve this, we fit an Akima spline to all sensor measurements and evaluate it at each environment timestep.
We chose the Akima interpolation method over a cubic spline, as it prevents the values from overshooting in-between sampled points.

\subsection{Distribution fitting}
Once a preprocessed dataset is available, we move on to the core part of DROPO---estimating the dynamics probability distribution $p_{\phi}(\xi)$.

In our method, the optimized parameter vector $\phi$ includes both the means and variances of each univariate distribution, assuming uncorrelated parameters. Therefore, with a dynamics parameter vector $\xi \in \mathbb R^d$, the parameterization is represented by a $2d$-dimensional vector $\phi \in \mathbb R^{2d}$. This is in contrast to DROID~\cite{tsai2021droid}, where only the means are optimized while the standard deviations are taken from the CMA-ES optimization distribution instead. This change allows to explicitly learn the variability of each simulated physical parameter to best explain heteroschedasticity and uncertainty in real-world data. DROPO can thus be thought as a random coefficient statistical model, where the optimized coefficients are themselves assumed to be random variables, as opposed to the more standard fixed-effects regression analysis. 

The distribution fitting process starts with an arbitrary initial guess $\phi_{init}=(\mu_{init}, \Sigma_{init})$ on the dynamics distribution, roughly informed by looking at the replayed trajectory in sim.
The main inference part follows by sampling $K$ dynamics parameters from $p_\phi(\xi)$.
Then, for each parameter vector $\xi_{k}$, we set the simulator state to the original real state $s_{t}$, execute the real action sequence $a_{t..t+\lambda-1}$ and observe the next state $s_{t+\lambda}^{\xi_{k}}$.
This process allows us to estimate the next-state distribution $p_{sim}(s^{sim}_{t+\lambda} | s_t, a_{t..t+\lambda-1}, \phi)$, modelled as a Gaussian distribution. In particular, we infer the corresponding mean and covariance matrix as follows:
\begin{equation} \label{eq:meanvar}
\begin{gathered} 
    \bar{s}^{\phi}_{t+\lambda} = \frac{1}{K} \sum_k s^{\xi_k}_{t+\lambda} \\
    \Sigma^{\phi}_{t+\lambda} = \Cov (p_{sim}(s^{sim}_{t+\lambda} | s_t, a_{t..t+\lambda-1}, \phi)) + \diag(\epsilon),
\end{gathered}
\end{equation}
where $\Cov(\cdot)$ is the unbiased sample covariance matrix estimator.
Here, $\epsilon$ is a hyperparameter used to compensate for observation noise and regularize the likelihood computation in case singular covariance matrices would appear---e.g. due to directions of variance in the state-space unexplainable by the simulator dynamics.
Tuning this hyperparameter will, effectively, adjust how much next-state variance is to be modelled by variance in dynamics parameters, relative to observation noise.
As demonstrated in our experiments, setting this parameter to too low values may result in obtaining an overly wide randomization distribution: in the extreme case of $\epsilon=0$, DROPO would attempt to model all phenomena that cannot be reproduced in a single simulation by variability in the dynamics, potentially overfitting to noise or unmodeled effects.
However, increasing $\epsilon$ to excessively large values will eventually lead to over-regularization and point-estimate dynamics, as all unexplained variance would be modeled as observation noise, falling back to pure system identification. To ensure stability while promoting variance in the dynamics parameters, we propose to tune $\epsilon$ by increasing it until the Mean Squared Error (MSE) between real and sim trajectories no longer meaningfully decreases---similarly to the Elbow method popularly used in machine learning.
We present a thorough analysis of the impact of the $\epsilon$ parameter in the Experiments section.

After observing the resulting simulator states, we make the assumption that the true, real-world observation $s_{t+\lambda}$ originates from the next-state distribution induced by dynamics randomization: $s_{t+\lambda} \sim \mathcal{N}(\bar{s}^{\phi}_{t+\lambda}, \Sigma^{\phi}_{t+\lambda})$

The log-likelihood of $s_{t+\lambda}$ under this distribution can be then calculated following the standard Gaussian formula (ignoring the constant term)
\begin{equation} \label{eq:loss}
    \mathcal{L}_{t} = \frac{1}{2} (\log \det \Sigma^{\phi}_{t+\lambda} + (\bar{s}^{\phi}_{t+\lambda} - s_{t+\lambda})^\top {\Sigma^{\phi}_{t+\lambda}}^{\hspace{-8pt}-1} (\bar{s}^{\phi}_{t+\lambda} - s_{t+\lambda}))
\end{equation}

The log-likelihoods of each individual transition can then be summed to obtain the total log-likelihood of the dataset $\mathcal{D}$ under $p_\phi(\xi)$:
    $\mathcal{L} = \sum_{t} \mathcal{L}_t$

We finally adjust the parameters $\phi$ to maximize the total log-likelihood $\mathcal{L}$.
As we don't make any assumptions about the differentiability of the simulator, we optimize $\mathcal{L}$ using a gradient-free optimization method.
DROPO can, in principle, work with an arbitrary optimization method; we use CMA-ES~\cite{hansen2001cmaes} throughout the experiments, with initial scale of 1 and optimization parameters normalized linearly (means) or in log scale (variances) to the interval $[0,4]$. Note that, in order to further improve stability, we model $p_{\phi}(\xi)$ as an uncorrelated truncated normal distribution bounded to values within two standard deviations away from the mean, with consecutive resampling when unfeasible parameters are drawn (e.g. negative masses).

\subsection{Policy training}
The policy $\pi(a|s)$ is trained for the given target task entirely in simulation using the converged domain randomization distribution $p_\phi(\xi)$. 
This is achieved by sampling new dynamics parameters $\xi \sim p_\phi(\xi)$ at the start of each training episode.
The RL policy can then be trained with an arbitrary reinforcement learning algorithm; we use Proximal Policy Optimization (PPO;~\cite{schulman2017proximal}) and Soft Actor-Critic (SAC;~\cite{Haarnoja18soft}) in our experiments.
The trajectories coming from the offline dataset are not used in any way during the policy training process; as such, the demonstrations are not required to be specific to the particular task, as long as they allow the relevant dynamics parameters to be identified.

\section{Experiments}
\label{sec:experiments}
The experimental evaluation aims to answer the following questions:

\begin{enumerate}
    \item is DROPO capable of recovering the original dynamics parameters and randomization distributions in simulated environments?;
    
    \item is DROPO capable of training a policy that can solve the task and transfer when unmodeled phenomena are present in simulation?;
    
    \item how does the value of the $\epsilon$ hyperparameter affect the obtained domain randomization distribution?;
    
    \item what is the real-world performance of policies trained with DROPO compared to DROID~\cite{tsai2021droid}, BayesSim~\cite{bayessimramos}, and uniform domain randomization (UDR)?

\end{enumerate}

\subsection{Benchmark methods}
Throughout our experiments we benchmark UDR similarly to~\cite{Muratore2021DataEfficientDR}, as we assess the average performance of several policies trained on different uniform bounds. More precisely, we train 10 policies with DR on 10 uniform bounds respectively, which are randomly sampled from the search spaces reported in the corresponding tables. This would reflect how manually picking uniform distributions within a reasonable search space affects the final performance.

Furthermore, in contrast to the original paper on DROID~\cite{tsai2021droid}, we replaced torque measurements with the raw state vectors when implementing the DROID baseline.
This modification makes more sense for the pushing and sliding tasks that we use for evaluation, as joint torques do not express enough information about the object being manipulated by the robot, especially for the sliding task---in contrast to the door opening task used in the original DROID paper.
The CMA-ES implementation used in DROPO and DROID is provided by Nevergrad~\cite{nevergrad}.

Finally, in order to adapt BayesSim to the offline setting with human demonstrations (where the data collection policy is not available), we repeat the original action sequence during data collection. Besides this modification, we make use of the original open-source repository provided by~\cite{bayessimramos}.
In most experiments, we evaluate both neural network (MDNN) and quasi-random Fourier (MDRFF) features and report the better result for brevity.
An exception to this is the sim-to-real experiment with the Panda robot, where results with both features are reported in order to compare the generalization capability of both variants.
Since BayesSim uses mixture Gaussian distributions, the distribution inference results reported in the tables and plots in this section are independent Gaussian approximations of the complete results, for the sake of brevity.
The complete BayesSim posteriors can be found in \ref{sec:bs:panda}.

\subsection{Hopper}
In the Hopper environment~\cite{openaigym}, a one-legged robot is tasked to learn to jump forward as fast as possible without falling down.
This environment has been used as a standard control benchmark in a variety of papers~\cite{schulman2017proximal,Haarnoja18soft,kumar19bear,rajeswaran_epopt_2017}.
We introduce parameter randomization to this environment by varying the mass of each link.
We use this environment as a simple benchmark to test DROPO's capability of identifying the dynamic parameters of the system and recovering the parameter distribution that was used to generate the dataset.
We later introduce an unmodeled effect in the form of a misspecified mass (which is not included in the optimization problem), similarly to~\cite{rajeswaran_epopt_2017}.
Given the simple sim-to-sim scenario, the offline trajectories for running DROPO have been collected by rolling out semi-converged policies trained directly in the target domain.

\subsubsection{Point-dynamics system identification} \label{sec:point_est_hopper}
When the offline dataset is collected on a single Hopper environment instance, DROPO is expected to identify the target physical parameters and to converge to point-estimate dynamics rather than distributions. We tested this behavior on the Hopper environment by collecting two trajectories (1000 state transitions corresponding to 8 seconds of wall time) on the ground truth dynamics shown in Table~\ref{tab:pointest_convergence}, both with and without noise of variance $10^{-5}$ injected in the offline observations.
The results are reported in the same table, as DROPO achieved convergence to the original dynamics regardless of the presence of noise and effectively decreased the standard deviation of learned parameters to $10^{-5}$, set as the lowest bound during the optimization problem.

In the point-estimate dynamics estimation, DROID also converged to the ground-truth parameter values, with a lower standard deviation than DROPO (as CMA-ES does not impose any lower bounds on the variance).
Thus, both DROPO and DROID perform virtually the same in this benchmark.

On the other hand, BayesSim overestimates the distribution variance and does not accurately recover the mean parameter value.
This is in line with the results reported in the original paper~\cite{bayessimramos}, where the resulting posterior distribution is much wider than the ground-truth values used for data generation.
While wider distributions may benefit generalization, BayesSim falls short of DROID and DROPO in terms of pure system identification capabilities.

\subsubsection{Recovering the dynamics distribution}
\label{sec:distribution_recovery}
In this experiment, the dataset $\mathcal{D}$ was collected while varying the target dynamics according to some ground truth distribution $p(\xi)$, which we want to recover. We collected two trajectories on the Hopper environment, varying all masses as $m_{i} \sim\mathcal{N}(\mu_{i}, 0.25^2)$ around their original values $\mu_{i}$ every 25 state transitions, except for the torso mass which was kept fixed. 
Since BayesSim expects the input to be trajectories collected from a single dynamics setting rather than a set of transitions, we had to use a different test procedure for this baseline; namely, we sample multiple dynamics conditions from the true distribution and run BayesSim individually for each of them.
The resulting distribution is then obtained by marginalizing the output distribution over the input dynamics distribution.

\begin{figure}
    \centering
    \includegraphics[width=0.98\linewidth]{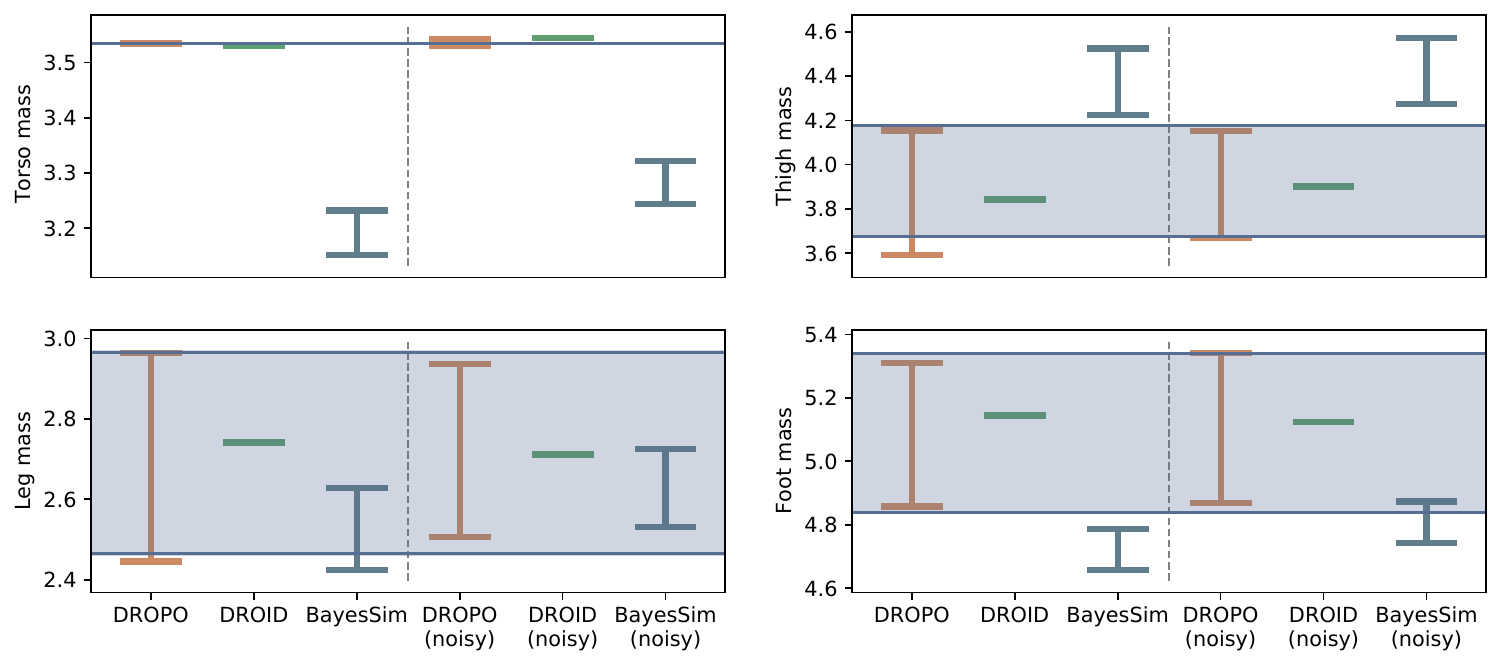}
    \caption{
    Dynamics distributions optimized by DROPO, DROID and BayesSim on offline data collected from a distribution of Hopper environments (blue shaded areas). The distributions are displayed with $2\sigma$-wide ranges centered on their means.
    }
    \label{fig:distr_recovery_results}
\end{figure}

The results are reported in Figure~\ref{fig:distr_recovery_results}, which shows DROPO successfully learning the original variance of the randomized masses, in addition to their means. We believe that such explainability is only possible when explicitly optimizing parameters through probabilistic distance metrics, as DROID's L2-based cost function fails at providing wide distribution ranges on the randomized masses. Note how BayesSim, while preserving some variance in the converged distributions, resulted in much less accurate dynamics distribution estimates. Analogously to the previous experiment, a noisy version of the dataset has also been tested, yielding similar results.

\subsubsection{Unmodeled phenomena}
Finally, we also tested DROPO's ability to compensate for unmodeled phenomena in simulation by assessing the complete transfer to a target simulator from an under-modeled source domain. We introduced such unmodeled effect by misspecifying the Hopper torso mass by 1kg in the source simulator and excluding it from the optimization space. Hence, each method may only optimize the means and variances of the remaining three masses to match the target transitions, and cannot adjust the incorrect torso mass.

To perform this evaluation, we first ran each method to infer the dynamics distribution given the same noiseless data as in Section~\ref{sec:point_est_hopper}.
Since there are no ground-truth domain parameter values that could be used as a reference when unmodeled phenomena are present, the domain randomization distributions cannot directly be used to quantitatively compare the estimated dynamics parameters.
Instead, we used the obtained distributions for policy training in the environment with a misspecified mass, and evaluated the resulting policies in the original environment where data was collected.
We repeated this process 3 times for each method, with different random seeds.
The policies were trained with SAC~\cite{Haarnoja18soft}.

\begin{figure}
    \centering
    \includegraphics[width=.7\linewidth]{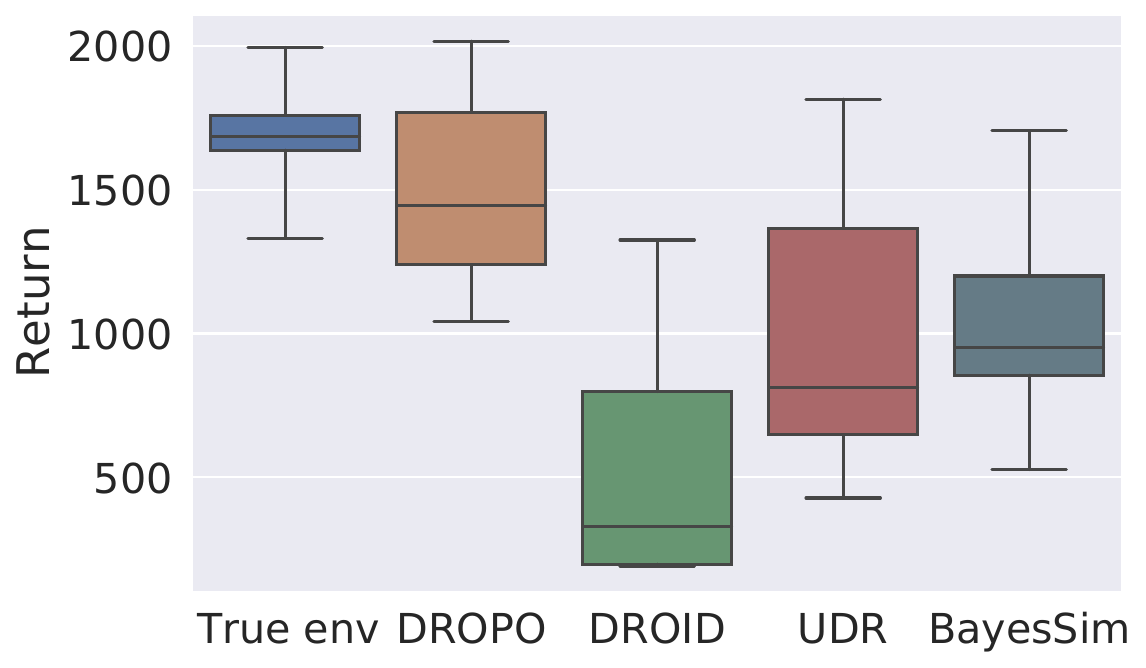}
    \caption{Sim-to-sim:~episode returns in the target Hopper environment, comparing policies trained directly in the target environment, and trained in the environment with a misspecified torso mass using domain randomization distributions inferred by DROID, DROPO and BayesSim.}
    \label{fig:hopper_fixedmass}
\end{figure}

The results of this evaluation are presented in Figure~\ref{fig:hopper_fixedmass}, together with a policy trained directly in the target environment to provide a notion of the upper bound.
We observe that policies trained with DROPO's randomization distributions outperform all baseline methods, almost matching the policy trained in the ground-truth environment in terms of performance.
At the same time, the policies obtained with DROID and BayesSim perform much worse.
Note how some UDR policies are at times able to generalize well to the target environment, suggesting that the task could be solved by manual trial and error with different uniform bounds.

To provide more insight into these results, we can look at the converged domain randomization distribution reported in Table~\ref{tab:pointest_convergence}.
Note how, even though the target dataset has been collected on point-dynamics parameters, DROPO converges to a distribution over the three randomized masses.
This behavior is indeed expected, as due to the unmodeled phenomenon introduced, a single point-estimate parameter vector is not able to accurately reproduce all transitions in the target dataset.
Therefore, DROPO widens the dynamics distribution to maximize the likelihood of target-domain data to be observed in simulation.
In addition, the optimized means are now much different from the ground truth parameters.
On the other hand, the results show that---just like in the distribution recovery experiments---DROID is eager to converge to point-estimate parameters, in order to minimize the L2-distance based objective function (see \ref{sec:appendix:least_squares_adr} for an intuitive explanation of this phenomenon).
Such a policy performs well in the training environment, but fails to generalize to the test conditions.

Furthermore, notice how BayesSim produced distributions with means located close the center of the search space with large standard deviations.
As such, it was able to learn a successful policy that works over a range of dynamics parameters, including the environment where the unmodeled phenomenon is present.
However, excessively large standard deviations result in overly conservative policies that produce a slower forward motion than DROPO, resulting in a lower return.
We suspect that this might be caused by the domain shift between the training data collected from the simulation with misspecified mass and the data from the real environment used to infer the randomization distribution.
In BayesSim, the inference network is trained on trajectories obtained from the simulator, each collected in different dynamics conditions.
As a result, any trajectory that cannot be reproduced by single dynamics conditions in simulation will effectively lie outside of the inference network's training domain.
Thus, we claim that any task involving unmodeled phenomena or unmodeled noise is substantially difficult for the BayesSim neural network, as the evaluation domain does not match the training domain anymore.
Later, we obtained similar results in the Panda sim-to-real experiments, as described in Section~\ref{sec:panda_real}.

To summarize, we observe that a combination of both system identification and domain randomization is essential to solve the task under this unmodeled effect, as neither UDR---pure unoptimized domain randomization---nor DROID---which converged to a point-estimate---result in a successful zero-shot transfer.

    \begin{table}[]
    \scriptsize
    \begin{tabular}{cllllll}
    \hline
                                                                        & \multicolumn{1}{c}{Masses (kg)}                                          &                       & $m_{1}$                                                                                         & $m_{2}$ & $m_{3}$ & $m_{4}$ \\ \hline
    \multicolumn{1}{l}{}                                                & \multicolumn{2}{l}{Ground truth}                                                                 & 3.534                                                                                            & 3.927    & 2.714    & 5.089    \\ \hline
    \multicolumn{1}{l}{}                                                &                                                                          & min                   & 1.767                                                                                            & 1.963    & 1.357    & 2.545    \\
    \multicolumn{1}{l}{\multirow{-2}{*}{}}                              & \multirow{-2}{*}{\begin{tabular}[c]{@{}l@{}}Search\\ space\end{tabular}} & max                   & 7.069                                                                                            & 7.854    & 5.429    & 10.18   \\ \hline
                                                                        &                                                                          & $\mu^{*}$                    & 3.534                                                                                            & 3.927    & 2.714    & 5.089    \\
                                                                        & \multirow{-2}{*}{\begin{tabular}[c]{@{}l@{}}DROPO\\ $^{\epsilon=1e-8}$\end{tabular}}                                                  & $\sigma^{*}$                     & 1.0e-5                                                                                           & 1.1e-5   & 1.1e-5  & 1.0e-5   \\ \cline{2-7}
                                                                        &                                                                          & $\mu^{*}$                     & 3.534                                                                                                & 3.927        & 2.714        & 5.089        \\
                                                                        & \multirow{-2}{*}{DROID}                                                  & $\sigma^{*}$                     & 1.2e-12                                                                                                & 2.1e-12        & 2.0e-12        & 1.1e-12        \\ \cline{2-7}
    \multirow{-6}{*}{Noiseless}                                         &                                                                          & $\mu^{*}$                     & 3.350                                                                                                & 5.342          & 3.994        & 5.560        \\
                                                                        & \multirow{-2}{*}{BayesSim}                                               & $\sigma^{*}$                  & 0.697                                                                                                & 0.707          & 0.686        & 0.541        \\ \hline
                                                                        &                                                                          & $\mu^{*}$                     & 3.531                                                                                            & 3.931    & 2.717    & 5.093    \\
                                                                        & \multirow{-2}{*}{\begin{tabular}[c]{@{}l@{}}DROPO\\ $^{\epsilon=1e-5}$\end{tabular}}                                                  & $\sigma^{*}$                     & 8.1e-3                                                                                           & 1.3e-3   & 1.3e-3   & 2.4e-4   \\  \cline{2-7}
                                                                        &                                                                          & $\mu^{*}$                     & 3.534                                                                                                & 3.927        & 2.714        & 5.089        \\
                                                                        & \multirow{-2}{*}{DROID}                                                  & $\sigma^{*}$                     & 1.2e-12                                                                                                & 2.4e-11        & 2.0e-12        & 1.2e-12        \\  \cline{2-7}
                                                                        &                                                                          & $\mu^{*}$                     & 6.295                                                                                                & 4.002          & 3.486        & 3.771        \\
    \multirow{-6}{*}{Noisy}                                             & \multirow{-2}{*}{BayesSim}                                               & $\sigma^{*}$                  & 0.218                                                                                                & 0.257          & 0.245        & 0.256        \\ \hline
                                                                        &                                                                          & $\mu^{*}$                   & {\color[HTML]{9A0000} }                                                                          & 4.707    & 3.181    & 4.385    \\
                                                                        & \multirow{-2}{*}{\begin{tabular}[c]{@{}l@{}}DROPO\\ $^{\epsilon=1e-3}$\end{tabular}} & $\sigma^{*}$    & \multirow{-2}{*}{{\color[HTML]{9A0000} \begin{tabular}[c]{@{}l@{}}2.534\\ (fixed)\end{tabular}}} & 0.667        & 0.206  & 0.826  \\  \cline{2-7}
                                                                        &                                                                          & $\mu^{*}$                   & {\color[HTML]{9A0000} }                                                                          & 4.021        & 3.591        & 4.845        \\
                                                                        & \multirow{-2}{*}{DROID}                                                  & $\sigma^{*}$    & \multirow{-2}{*}{{\color[HTML]{9A0000} \begin{tabular}[c]{@{}l@{}}2.534\\ (fixed)\end{tabular}}} & 6.0e-6        & 1.7e-6        & 4.2e-6       \\  \cline{2-7}
                                                                        &                                                                          & $\mu^{*}$                   & {\color[HTML]{9A0000} }                                                                          & 3.493      & 2.901    & 4.596     \\
    \multirow{-6}{*}{Unmodeled}                                        & \multirow{-2}{*}{BayesSim}                                               & $\sigma^{*}$                & \multirow{-2}{*}{{\color[HTML]{9A0000} \begin{tabular}[c]{@{}l@{}}2.534\\ (fixed)\end{tabular}}} & 1.576      & 1.471    & 1.537     \\

    \hline
    \end{tabular}
    \caption{Optimized dynamics distributions on Hopper OpenAI Gym environment. The ground truth values indicate the simulator instance used during data collection.}
    \label{tab:pointest_convergence}
    \end{table}

\subsubsection{The domain variance–observation noise tradeoff}
Throughout our experiments, we observed that introducing the $\epsilon$ hyperparameter is generally beneficial to obtain more stable and sound results (see Section~\ref{sec:method}). This turned out to be particularly effective when optimizing the estimated likelihood function in presence of noise or unmodeled effects. However, overly large values of $\epsilon$ should be avoided to prevent DROPO from converging to point-estimate dynamics parameters.
To study this trade-off, we report a thorough sensitivity analysis on this hyperparameter, showing its impact on the final performance of DROPO for the previously described Hopper experiments and describing our tuning procedure in more detail.

\begin{figure}
    \centering
    \begin{subfigure}{0.6\linewidth}
    \centering
    \includegraphics[width=\linewidth]{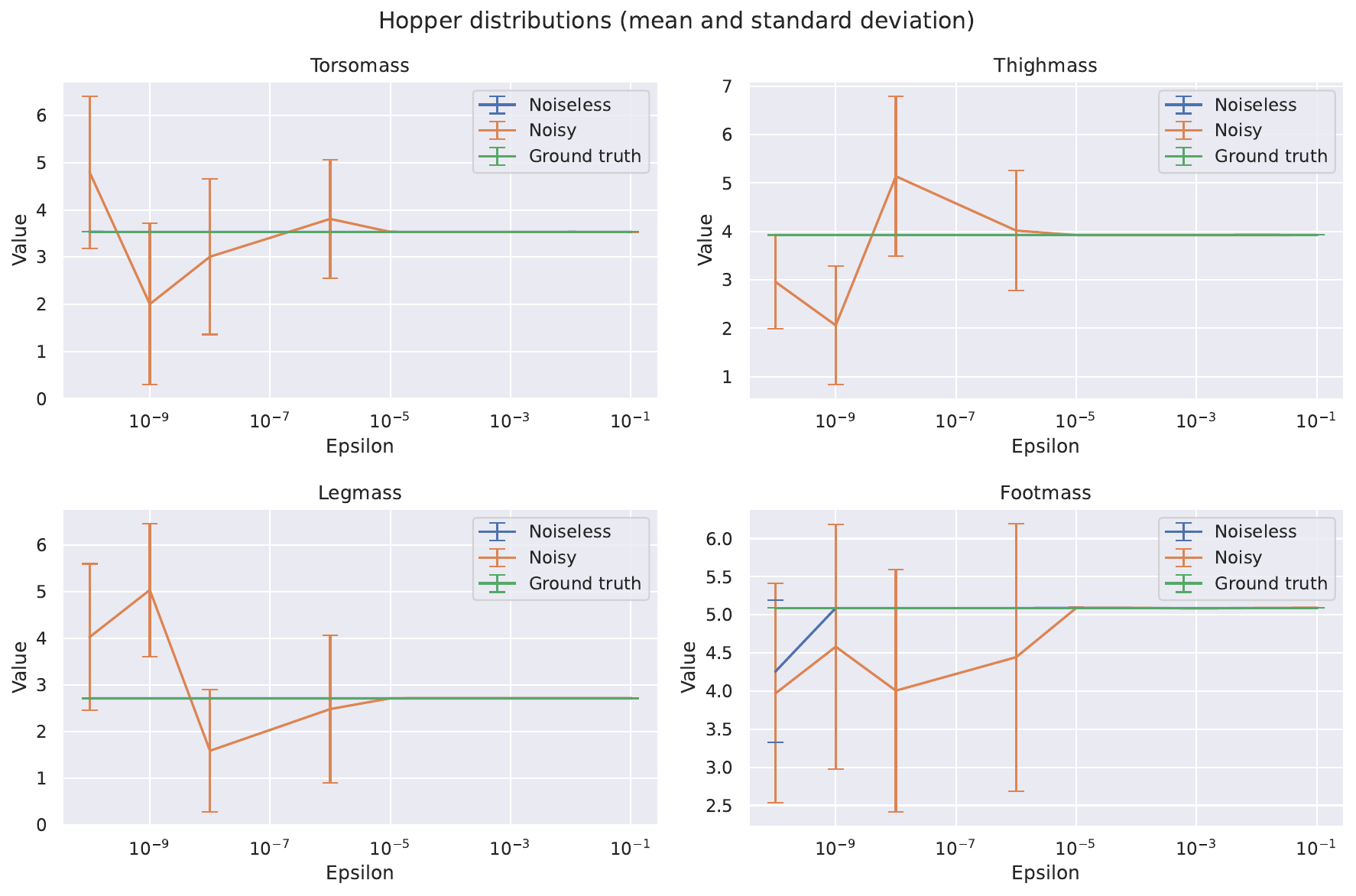}
    \caption{}
    \label{fig:hop_pointest_dists}
    \end{subfigure}
    \begin{subfigure}{0.38\linewidth}
    \centering
    \includegraphics[width=\linewidth]{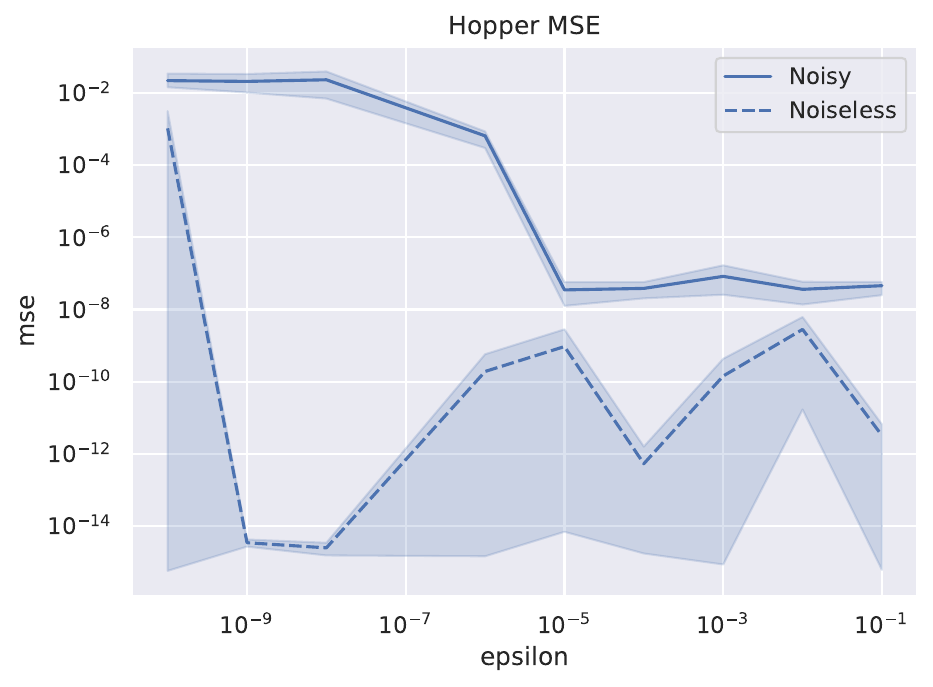}
    \caption{}
    \label{fig:hop_pointest_mse}
    \end{subfigure}
    \caption{Epsilon sweep on point-estimate dynamics recovery: (\subref{fig:hop_pointest_dists}) shows the converged distributions in the noiseless and noisy ($\sigma^2 = 10^{-5}$) cases, (\subref{fig:hop_pointest_mse}) shows how the MSE changes with $\epsilon$. (results averaged over 4 random seeds)}
    \label{fig:epsilon_sweep_pointest}
\end{figure}

The results of point-dynamics recovery in Hopper with varying $\epsilon$ are presented in Figure~\ref{fig:epsilon_sweep_pointest}. 
We observe that, for a low value of $\epsilon$, DROPO overestimates the standard deviation in the noisy setting, as it attempts to explain the random variation in next-states by variance in dynamics. 
However, for $\epsilon$ values equal to the noise variance (or higher), DROPO converges to very low standard deviations, corresponding to a point-estimate of dynamics parameters and leading to a significant drop in MSE. The MSE is computed w.r.t. the converged distribution $\phi^{*}$ obtained by DROPO, as follows: $MSE(\phi^{*})=\sum_{t} \|(\bar{s}^{\phi^*}_{t+1} - s_{t+1})\|^{2}$. This metric resembles the trajectory discrepancy function adopted by previous works~\cite{chebotar19closing, tsai2021droid}, and we propose to use it as a reference to reach a stable optimization process and account for observation noise when tuning $\epsilon$. Note how, in the same figure, the MSE drop is not as clear for the noiseless setting: here, $\epsilon$ only serves the purpose to regularize the log-likelihood computation, which could be ill-conditioned with too low values (e.g. $\epsilon=10^{-10}$ in Figure~\ref{fig:hop_pointest_mse}) as close-to-singular covariance matrices may appear when estimating the next-state distribution\footnote{Adding a small value $\epsilon$ to the diagonal of an ill-conditioned covariance matrix has the effect to increase all eigenvalues by $\epsilon$, thus making the likelihood computation more stable.}.

\begin{figure}
    \centering
    \begin{subfigure}{0.6\linewidth}
    \centering
    \includegraphics[width=\linewidth]{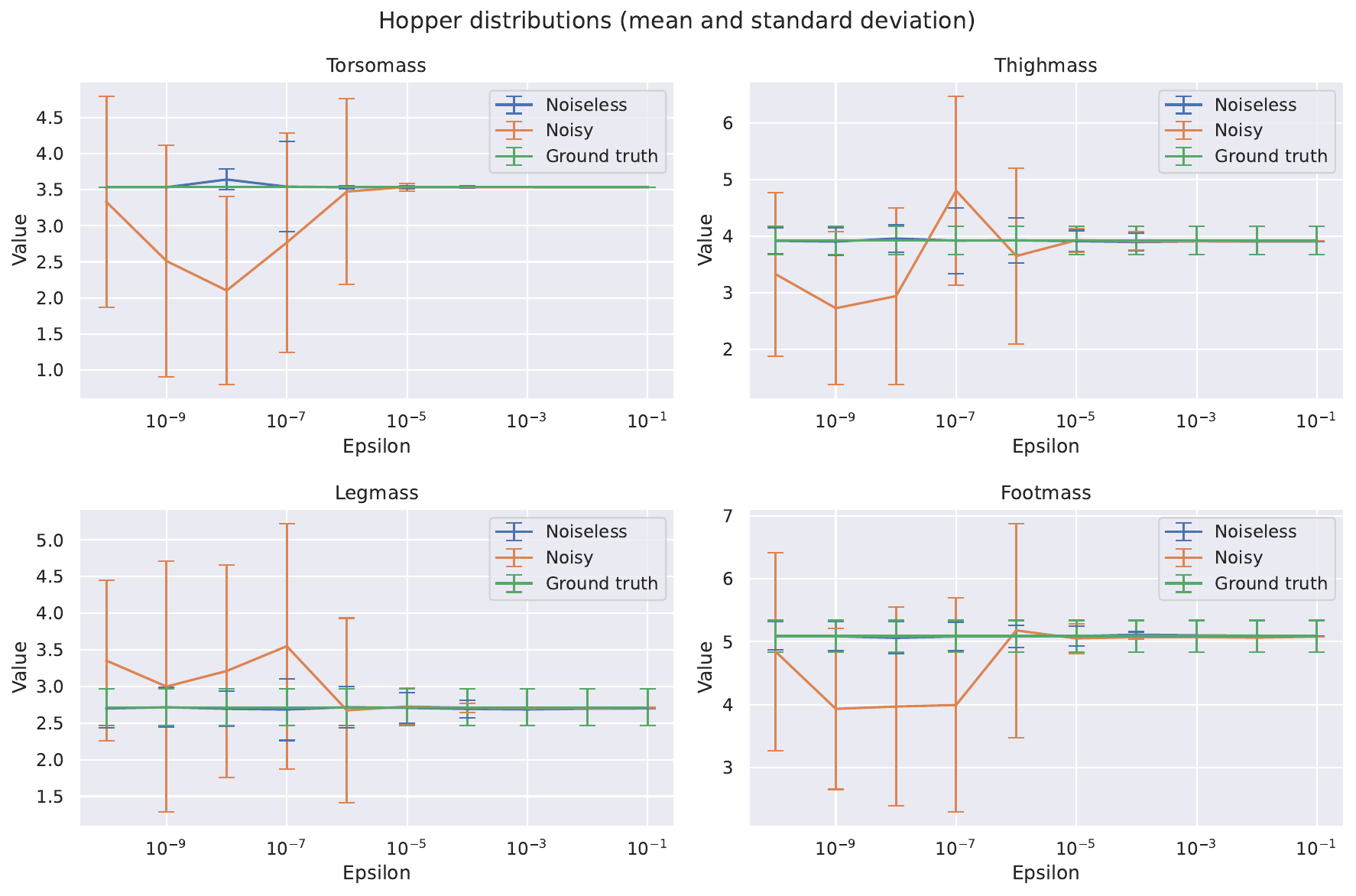}
    \caption{}
    \label{fig:hop_distrecv_dists}
    \end{subfigure}
    \begin{subfigure}{0.38\linewidth}
    \centering
    \includegraphics[width=\linewidth]{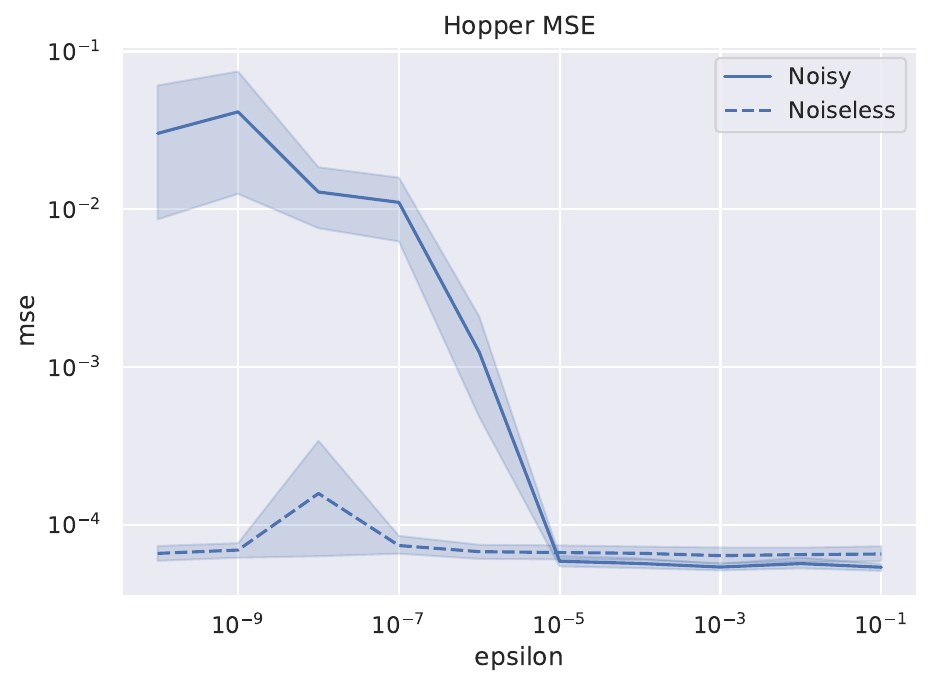}
    \caption{}
    \label{fig:hop_distrecv_mse}
    \end{subfigure}
    \caption{Epsilon sweep on dynamics distribution recovery: (\subref{fig:hop_distrecv_dists}) shows the converged distributions in the noiseless and noisy ($\sigma^2 = 10^{-5}$) cases, (\subref{fig:hop_distrecv_mse}) shows how the MSE changes with $\epsilon$. (results averaged over 4 random seeds)}
    \label{fig:epsilon_sweep_distrrec}
\end{figure}

We observe a similar phenomenon in the dynamics distribution recovery scenario (Figure~\ref{fig:epsilon_sweep_distrrec}).
Small $\epsilon$ values result in an excessively large variance in masses in presence of noise, while correctly identifying the ground-truth distribution in the noiseless case.
As $\epsilon$ grows larger, more of the variety in next-states gets attributed to noise and the estimated dynamics distributions move towards point dynamics, both in the noisy and noiseless case.
The best ground-truth distribution recovery is observed when $\epsilon$ matches the true noise variance, which also corresponds to the sharp drop in MSE observed in Figure~\ref{fig:hop_distrecv_mse}---thus validating the proposed tuning process.

\begin{figure}
    \centering
    \begin{subfigure}{0.6\linewidth}
    \centering
    \includegraphics[width=\linewidth]{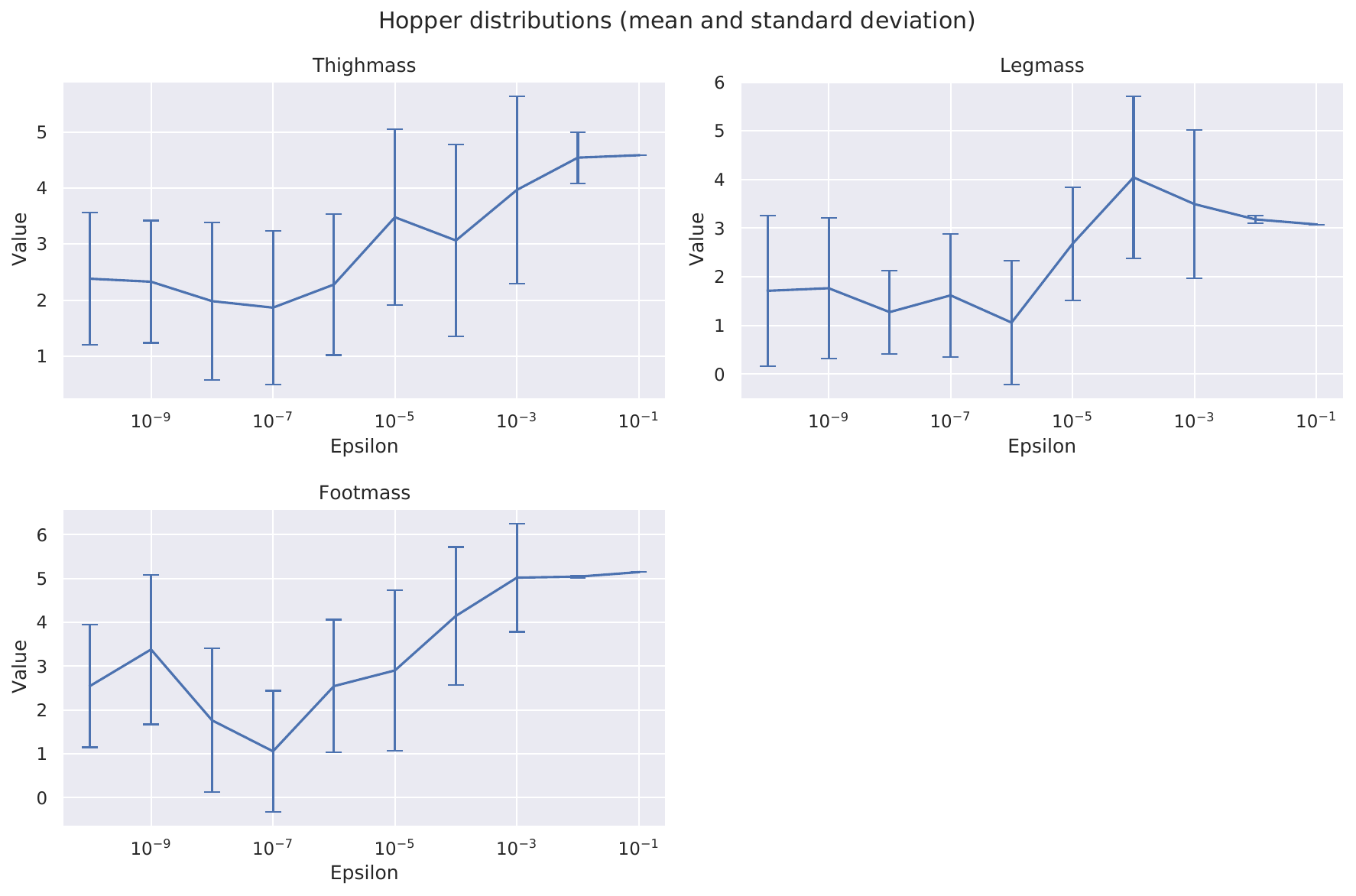}
    \caption{}
    \label{fig:hop_wrongm_dists}
    \end{subfigure}
    \begin{subfigure}{0.38\linewidth}
    \centering
    \includegraphics[width=\linewidth]{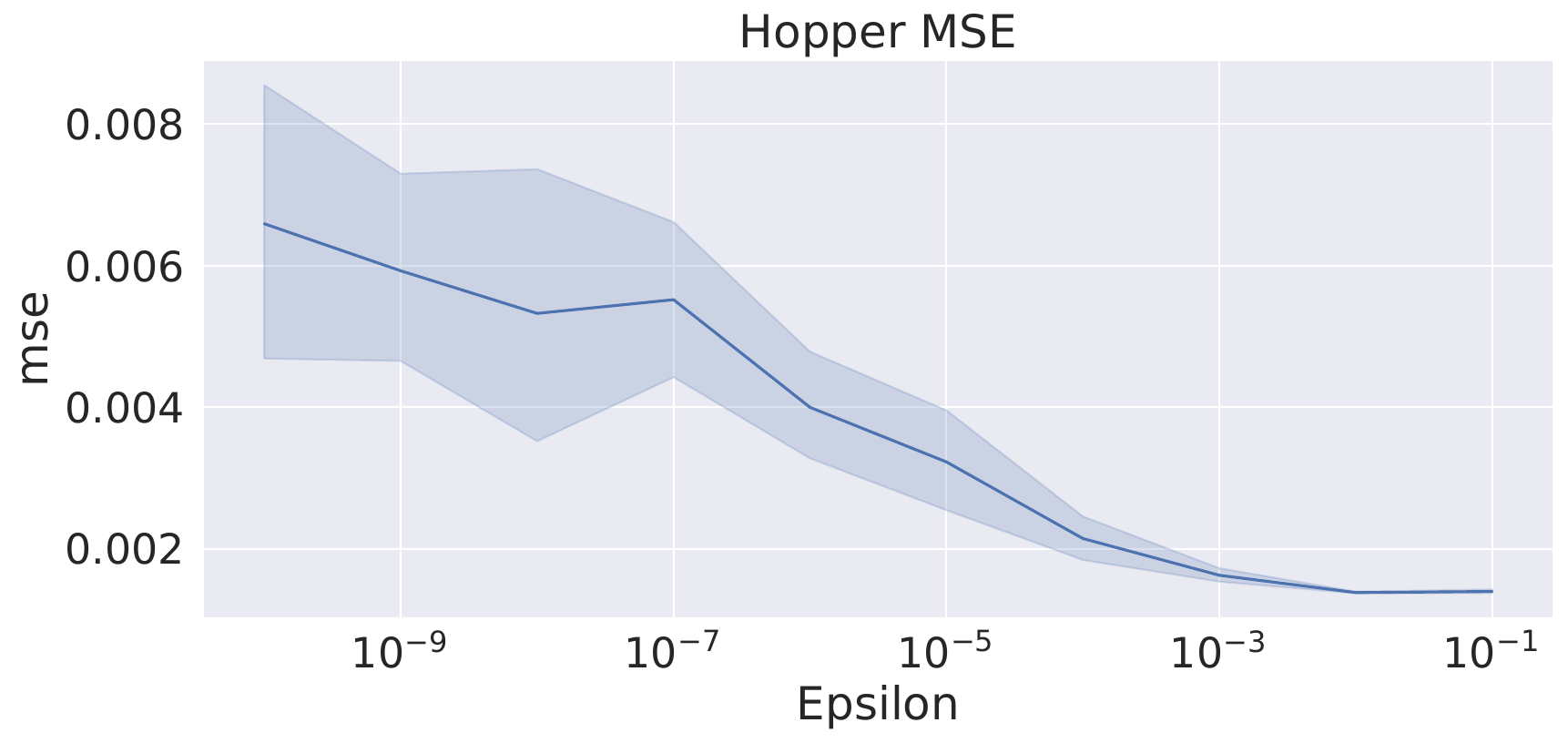}
    \caption{}
    \label{fig:hop_wrongm_mse}
    \end{subfigure}
    \caption{Epsilon sweep on the unmodeled phenomenon experiment: (\subref{fig:hop_wrongm_dists}) shows the converged distributions, (\subref{fig:hop_wrongm_mse}) shows how the MSE changes with $\epsilon$. (means and variances averaged over 8 seeds)}
    \label{fig:epsilon_sweep_unmodeled}
\end{figure}

Finally, we report similar observations for the unmodeled phenomenon experiment (Figure~\ref{fig:epsilon_sweep_unmodeled}), where---as $\epsilon$ increases---unexplainable effects in state space due to under-modeled dynamics get attributed to noise, stabilizing the optimization problem.
Once again, we tuned $\epsilon$ based on the MSE observed in Figure~\ref{fig:hop_wrongm_mse}, and picked $\epsilon=10^{-3}$, which is where the MSE values stop to meaningfully improve.
As observed in Figure~\ref{fig:hop_wrongm_dists}, this value closely approximates the mean obtained with higher $\epsilon$ values, while preventing the distribution from collapsing due to excessive regularization.
To this end, we would like to emphasize that a conservative choice may always be made with DROPO when tuning the value of $\epsilon$: higher values will only shift DROPO towards lower standard deviations (eventually providing similar results to DROID), while allowing a desirably low MSE to be obtained. 

Note how the proposed tuning procedure draws similarities to the elbow method popularly used in machine learning, e.g. to find the best number of clusters in clustering analysis or the number of principle components to retain in principal component analysis.

\begin{figure}
    \centering
    \includegraphics[width=0.7\linewidth]{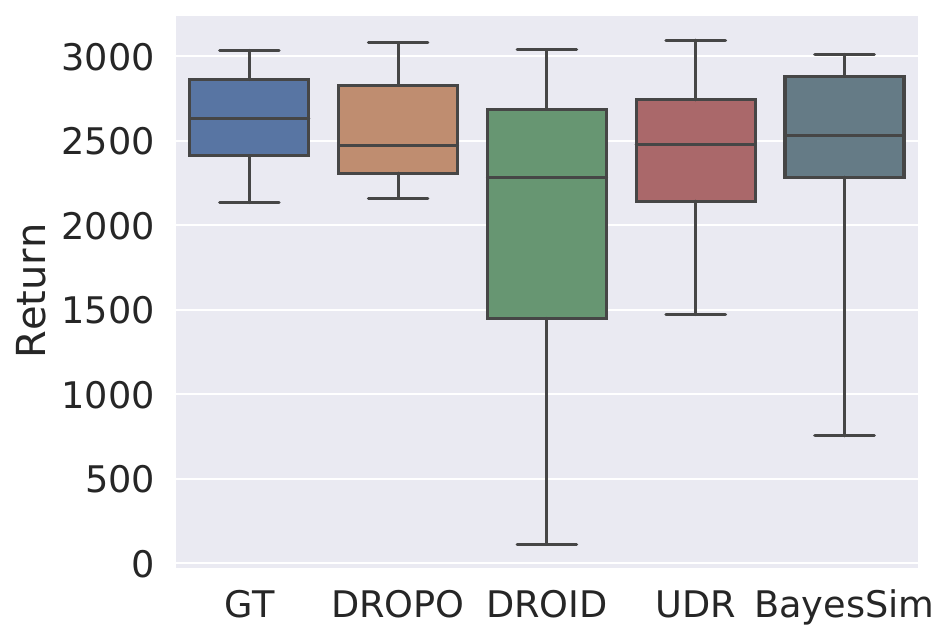}
    \caption{Sim-to-sim:~episode returns in the target simulated environment when trained on domain randomization distributions inferred by DROID, DROPO and BayesSim and on the ground-truth dynamics distribution (GT) in the Panda pushing task.}
    \label{fig:push_sim2sim}
\end{figure}

\subsection{Robotics experiments}
We further evaluate DROPO on two sim-to-real robotics setups: a hockey puck sliding and a box pushing task.

\subsubsection{HockeyPuck}
\begin{figure}
    \centering
    \begin{subfigure}{0.429\linewidth}
        \centering
        \includegraphics[width=\linewidth]{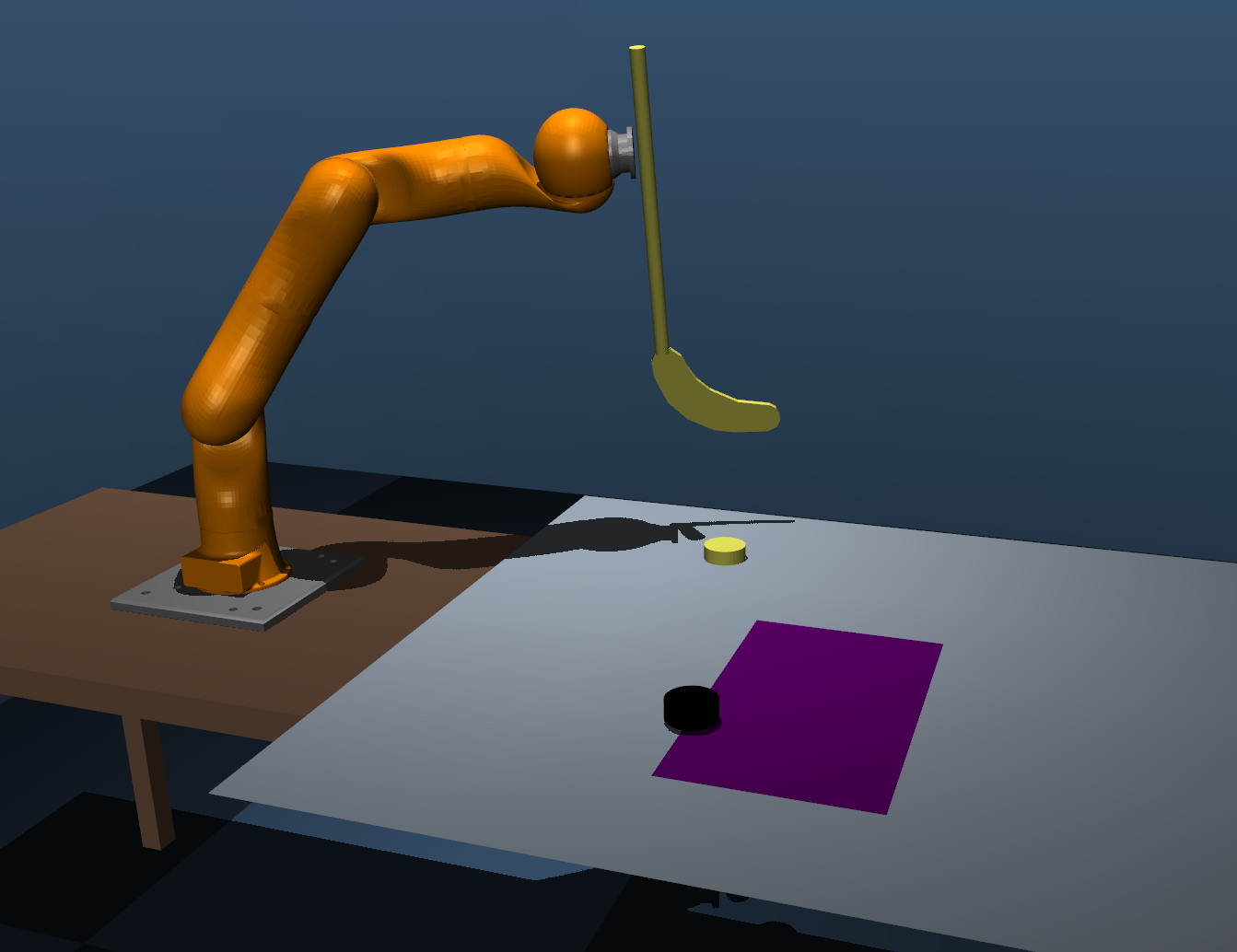}
        \caption{}
        \label{fig:hpuck_sim}
    \end{subfigure}
    \begin{subfigure}{0.495\linewidth}
        \centering
        \includegraphics[width=\linewidth]{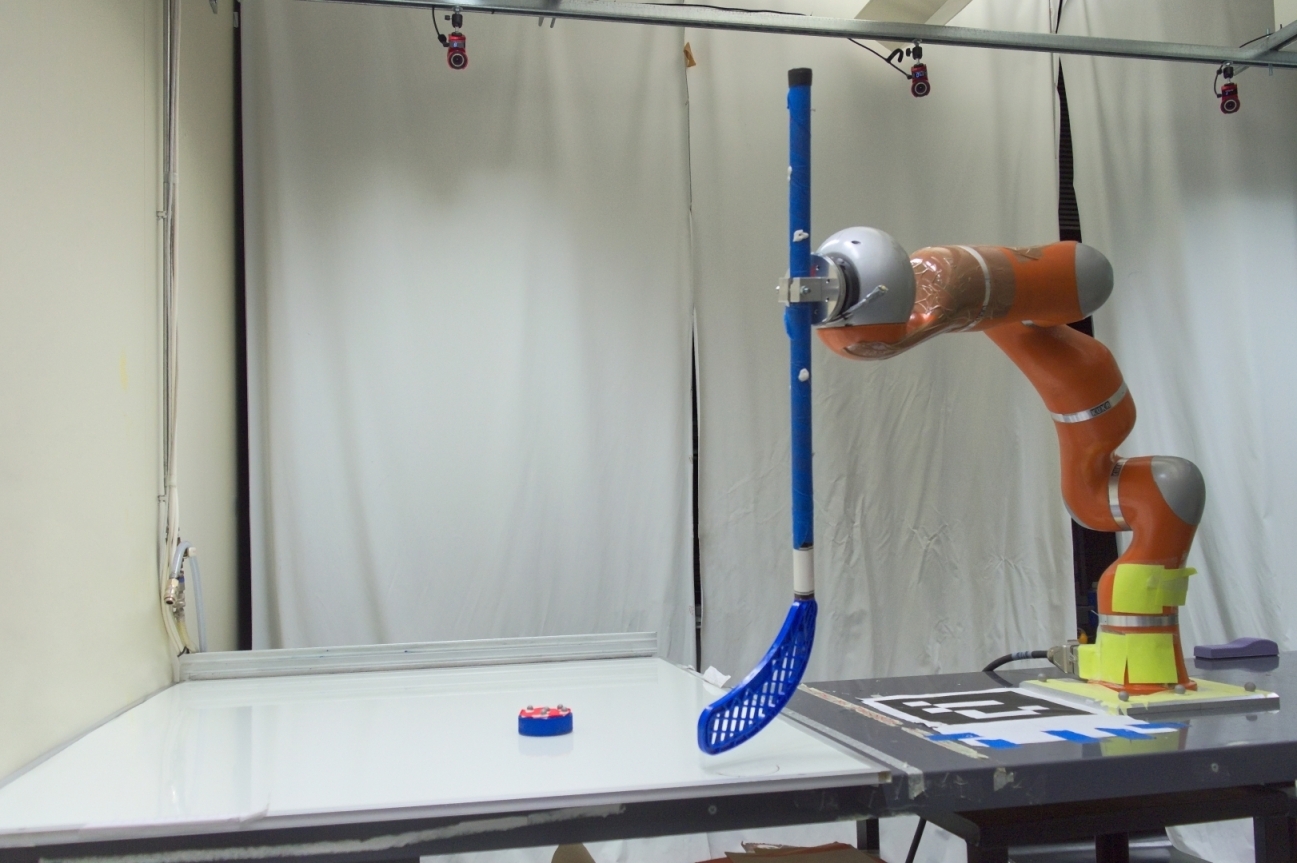}
        \caption{}
        \label{fig:hpuck_real}
    \end{subfigure}
    \caption{The hockeypuck setup in simulation~(\subref{fig:hpuck_sim}) and in the real-world~(\subref{fig:hpuck_real}). The purple area in~(\subref{fig:hpuck_sim}) shows the range of possible goal positions.}
    \label{fig:hpuck}
\end{figure}
In the HockeyPuck setup, the Kuka LWR4+ is equipped with a hockey stick and tasked with hitting a hockey puck such that it stops at the given target location.
For this setup, we use an ice hockey puck and a whiteboard as a low-friction surface for the puck to slide on.
The simulation environment is built in MuJoCo~\cite{todorov2012mujoco} akin to the real-world setup.
Both setups are shown in Figure~\ref{fig:hpuck}.
The yellow puck and the purple area in Figure~\ref{fig:hpuck_sim} show the initial position of the puck and the range of possible goal positions.

In this setup, the actions are whole hitting trajectories; this is achieved by first training a variational autoencoder~(VAE) on a range of task-specific trajectories, and later using the decoder as a trajectory generator, with actions given in the latent space of the VAE, following~\cite{ghadirzadeh2017deep,hamalainen2019affordance}.
This simplifies the reinforcement learning problem by turning it into a contextual multi-armed bandit problem.

The demonstrations for optimizing the parameter distribution are obtained by rolling out 5 random trajectories of 750 transitions (or 7.5 seconds) each, sampled from the latent space of the VAE ($z \sim \mathcal{N}(0, I)$).
The joint positions are obtained using the robot's internal sensors, while the positions of the hockey puck are obtained with OptiTrack, a motion capture system, with the frequency of 120Hz.
During the post-processing, the data is synchronized, the positions are resampled to match the simulated timesteps of the simulated environment, and the velocities are obtained by taking the derivative of the spline used to resample the signal.

The state space used for dynamics fitting with DROPO contains the positions and velocities of the robot joints and the hockey puck.
The use of a trajectory generator allows us to use original commanded velocities as actions when replaying trajectories in the simulator, and---on this particular setup---removes the need of action inference from demonstrations.

We randomize a total of 18 dynamics parameters: the mass of the puck, the puck-surface friction coefficients, the time constant of the stick-puck contact---regulating the stiffness and damping while keeping the interaction critically damped---and the gains of the position controller used in simulation. Note that, while more parameters may be added to the optimization problem to increase the model's explanatory power, complex correlations between the parameters may arise in addition to a generally higher complexity.

We use the reward function of~\cite{Levine16endtoend}: $r = -d^2 - log(d + \alpha)$,
where $d$ is the distance to the target and $\alpha$ is a constant (we use $\alpha=10^{-3}$ throughout the experiments).

\begin{figure}
    \centering
    \includegraphics[width=.6\linewidth]{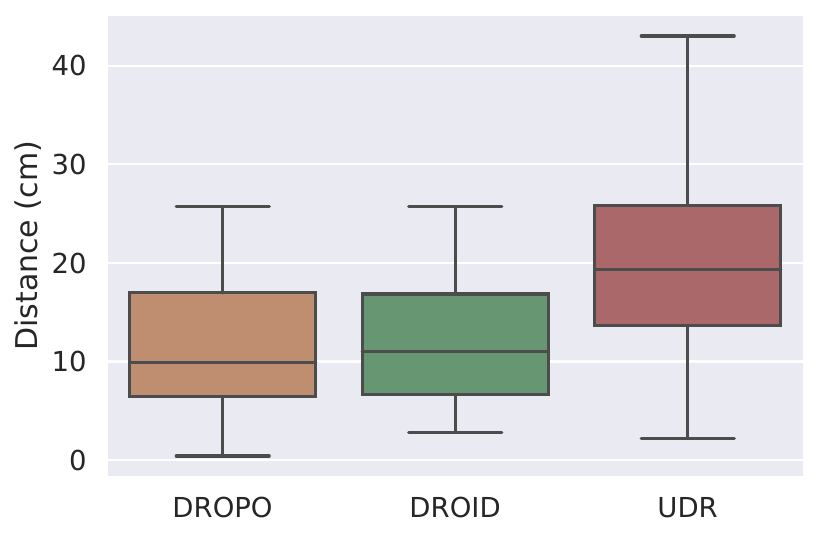}
    \caption{Sim-to-real performance in terms of the final distance to the target position on the hockeypuck setup (in centimeters; lower is better).}
    \label{fig:hp_simtoreal}
\end{figure}

The results comparing DROPO, DROID and UDR are shown in Figure~\ref{fig:hp_simtoreal}.
We see that, on this setup, both DROPO and DROID significantly outperformed UDR, and that there was no significant difference between the two methods, despite DROID essentially converging to point estimates (see Table~\ref{tab:sim2real_hockeypuck}).
This is likely caused by the bandit formulation and the resulting feed-forward nature of the task; once the action is selected, the agent does not receive any feedback from the environment.
Under this setup, it is thus optimal for the policy to select the action (or the trajectory) which minimizes the expected final distance from the target position under the dynamics distribution used in training.
This closely resembles the optimization criterion used by DROID (minimizing the expected L2 norm between trajectories).
Additionally, under this formulation the policy lacks the ability to make up for possible deviations from the trajectory obtained in point-estimate dynamics.
In a sequential decision making problem with uncertain dynamics, the agent needs to be able to recover from a wider range of states, due to the uncertainty in state transitions that stems from the uncertainty in dynamics.
In a bandit problem, however, the system operates in a feedforward manner, and thus is not able to handle such situations at all.
As a result, the benefits of having a distribution of dynamics are diminished.

\begin{table}[]
\scriptsize
\centering
\begin{tabular}{llrrrr}
\hline
\multicolumn{1}{c}{}                                                                &              & \multicolumn{1}{c}{$m$} & \multicolumn{1}{c}{$f_{x}$} & \multicolumn{1}{c}{$f_{y}$} & \multicolumn{1}{c}{$t_{const}$} \\ \hline
\multirow{2}{*}{\begin{tabular}[c]{@{}l@{}}Search\\ space\end{tabular}}             & min          & 0.08                    & 0.2                         & 0.2                         & 0.001                           \\
                                                                                    & max          & 0.2                     & 0.75                        & 0.75                        & 0.02                            \\ \hline
\multirow{2}{*}{\begin{tabular}[c]{@{}l@{}}DROPO\\ $^{\epsilon=1e-2}$\end{tabular}} & $\mu^{*}$    & 0.111                   & 0.349                       & 0.231                       & 0.019                           \\
                                                                                    & $\sigma^{*}$ & 2.2e-04                 & 3.4e-05                     & 4.7e-04                     & 6.7e-05                         \\
\multirow{2}{*}{DROID}                                                              & $\mu^{*}$    & 0.100                   & 0.334                       & 0.226                       & 0.019                           \\
                                                                                    & $\sigma^{*}$ & 1.5e-07                 & 8.9e-06                     & 9.7e-06                     & 1.7e-05                         \\ \hline
\end{tabular}
\caption{Optimized dynamics distributions on 5 trajectories collected on the Hockeypuck environment. The converged mass $m$, friction coefficients $f_{x}$ and $f_{y}$ along the two axes, and timeconst parameter $t_{const}$ are reported.}
\label{tab:sim2real_hockeypuck}
\end{table}

\subsubsection{Pushing} \label{sec:panda_real}
\begin{figure}
    \centering
    \begin{subfigure}{0.312\linewidth}
        \centering
        \includegraphics[width=\linewidth]{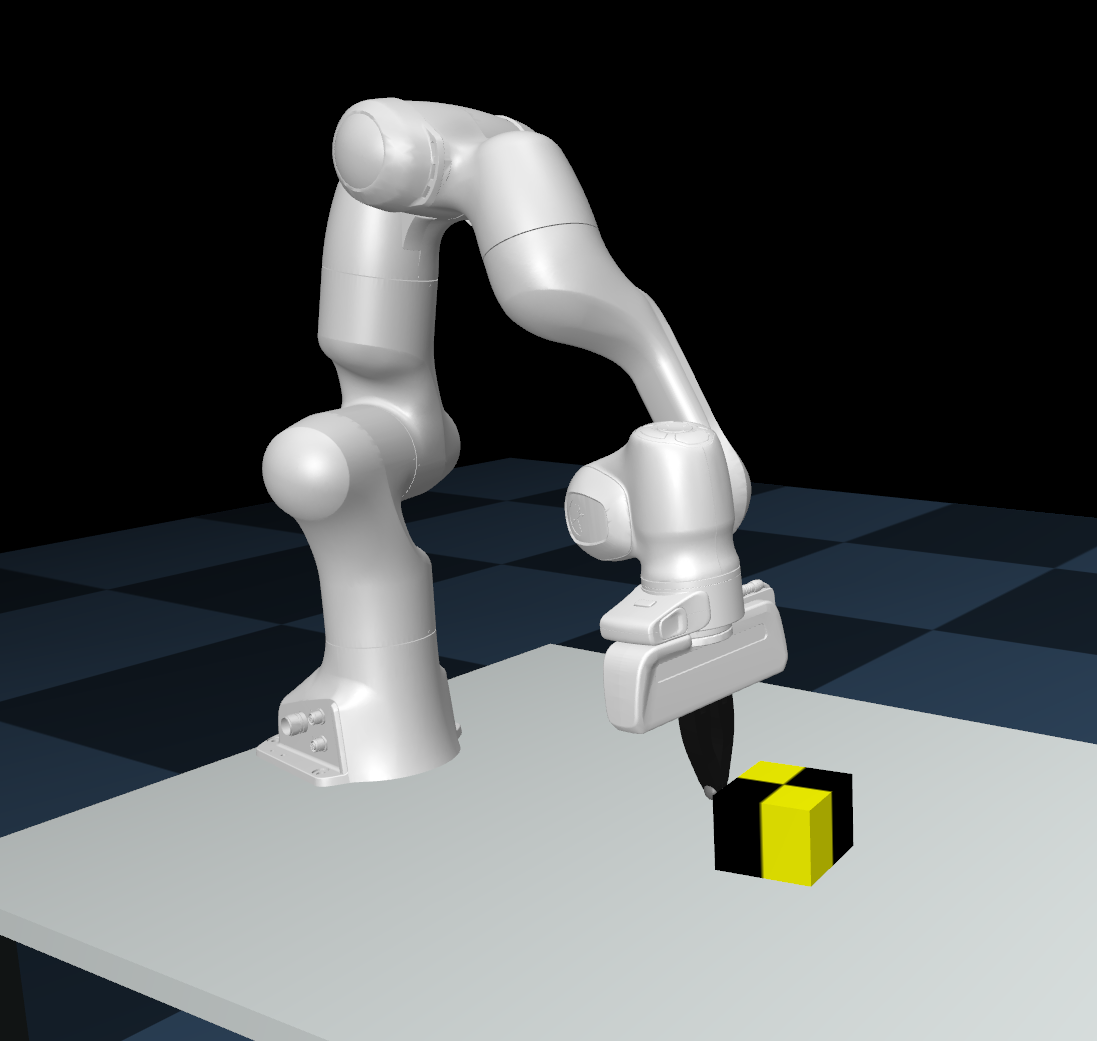}
        \caption{}
        \label{fig:push_sim}
    \end{subfigure}
    \begin{subfigure}{0.36\linewidth}
        \centering
        \includegraphics[width=\linewidth]{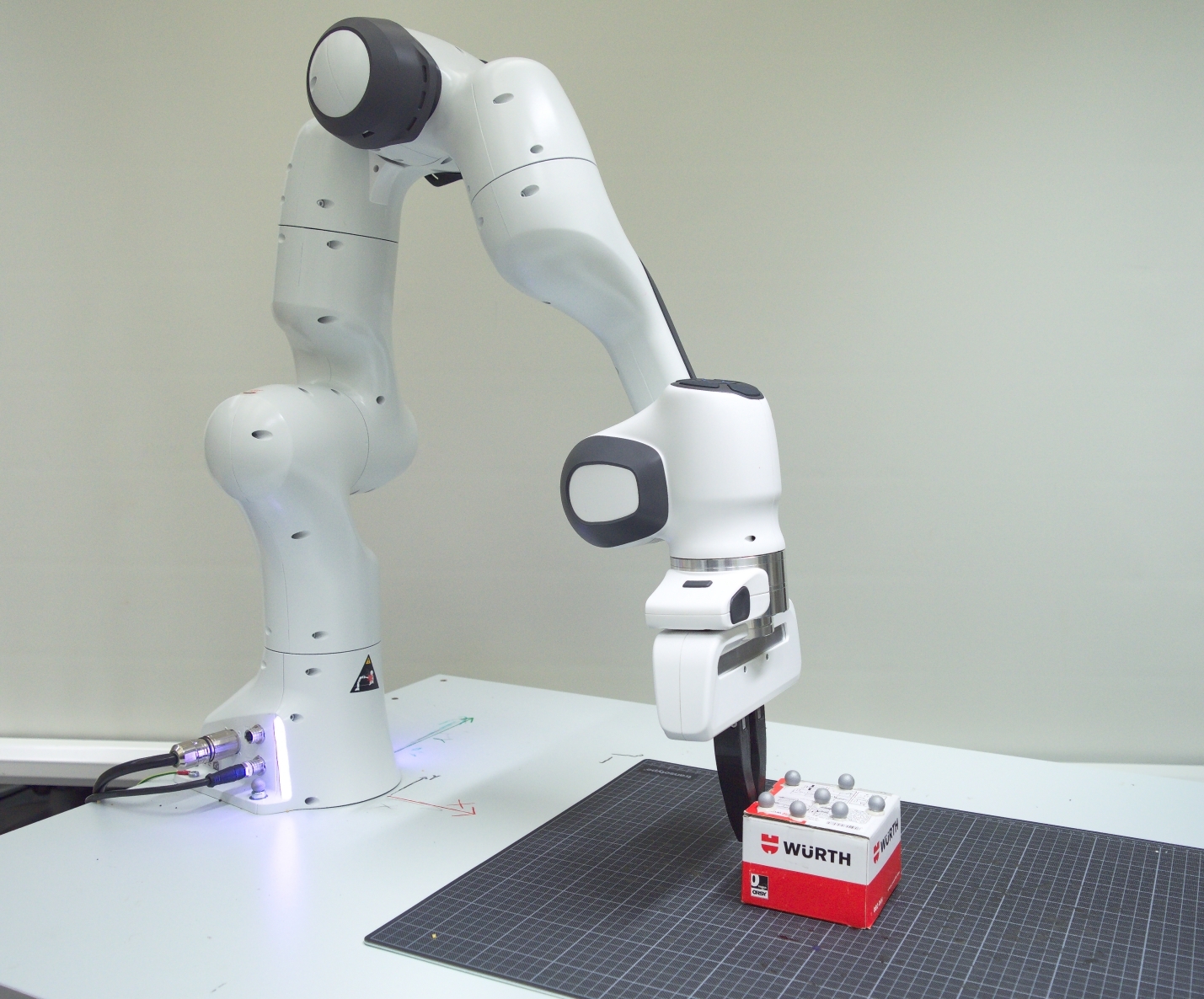}
        \caption{}
        \label{fig:push_real}
    \end{subfigure}
    \begin{subfigure}{0.25\linewidth}
        \centering
        \includegraphics[width=\linewidth]{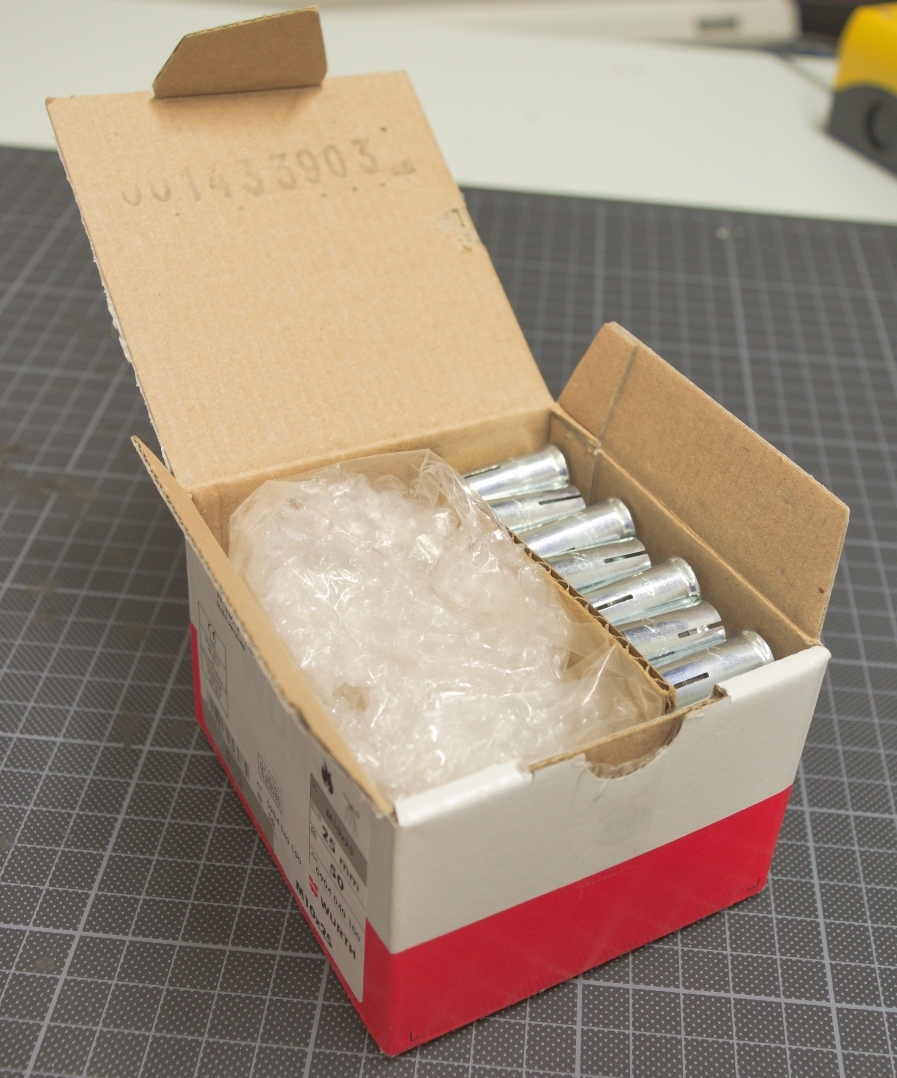}
        \caption{}
        \label{fig:push_box}
    \end{subfigure}
    \caption{The push setup in simulation~(\subref{fig:push_sim}) and in the real world~(\subref{fig:push_real}), with the insides of the box shown in~(\subref{fig:push_box}).}
    \label{fig:push_setup}
\end{figure}
We finally evaluated DROPO on a pushing task with the Franka Panda robot.
In this setup, shown in Figure~\ref{fig:push_setup}, the goal of the robot was to push a box to a fixed target position.
To make this task more challenging, the box center of mass was altered by filling it up with heavy steel bolts on one side and bubble wrap on the other, as shown in Figure~\ref{fig:push_box}. Therefore, we randomize a total of 5 dynamics parameters to shorten the reality gap in this setup: mass of the box, box-table friction coefficients $f_x$ and $f_y$, and center of mass offset from the geometric center ($com_x$, $com_y$). Due to the offset center of mass, we additionally included the orientations and angular velocity of the box in the state space used for fitting DROPO.

In order to model log-likelihood of orientations with DROPO, we used quaternion representations and evaluated the angle between the quaternion in the real-world dataset and the one in simulation; this angle, $\alpha$, then replaced the orientation inside the next state $s^\xi_{t+1}$, with $\alpha$ inside the real-world state vector set to $0$.
Since such a distribution would not be well-modeled with a Gaussian, for each transition we insert two resulting next states (with $\alpha$ and $-\alpha$), resulting in a symmetric distribution of angles centered around $0$.

We first tested our method in a sim-to-sim version of the task: synthetic data was collected by a forward-pushing policy as demonstrations for DROPO, DROID and BayesSim, sampling new dynamics parameters from the ground-truth distribution after every rollout, for a total of 50 trajectories of 30 transitions each, amounting to 30 seconds of wall-time interaction.
The resulting DR distributions are presented in Table~\ref{tab:sim2sim_push_bounds}.
Like before, the reported BayesSim results are independent Gaussian approximations of the actual results, which can be found in~\ref{sec:bs:panda}.
We again observe that DROID converged to a point estimate, with standard deviations in the order of $10^{-10}$--$10^{-11}$, while DROPO results in a distribution that is relatively close to the ground truth, albeit not the same.
The largest discrepancy can be observed in the inferred friction value along the  $x$ direction.
Since the trajectory used for this demonstration was pushing the box primarily away from the robot (along the $y$ axis), accurately inferring friction along the lateral axis is challenging, and the variations in box motion resulting from variations in this friction are minor and can be reasonably well modeled by variation in other parameters.
Similarly, the mass of the box can also be relatively difficult to identify, given that the inference is performed with a position controller and the mass of the box is much smaller than the mass of the robot.

We then used these DR distributions to train a policy for pushing the box to a goal on the front-left side of the robot.
We used the same reward function as for the Hockey experiments, with additional control penalties added to regularize the motions and to deter the simulated robot from going over the velocity and acceleration limits of the physical robot arm.
The performance of each policy is shown in Figure~\ref{fig:push_sim2sim}.
We observe that, despite the converged distribution being slightly different than the ground truth, the policies trained with DROPO performed close to the ground truth policies, while DROID policies performed noticeably worse, likely due to overfitting to point-estimate dynamics.
Interestingly, the UDR baseline outperformed DROID too. This suggests that, for the underlying task, DR is a more effective transfer technique than system identification even when the distribution is not carefully tuned; in other words, wider but less accurate dynamics parameter distributions provide better results than converged point estimate parameter values.

We then evaluated DROPO's ability to transfer to the real-world setup, in comparison to DROID, BayesSim, and UDR.
A single task-agnostic trajectory of about 13 seconds was collected through kinesthetic guidance and preprocessed in a similar way to hockey puck---we used robot's joint encoders to get the joint positions and OptiTrack to track the object.

\begin{figure}
    \centering
    \includegraphics[width=.8\linewidth]{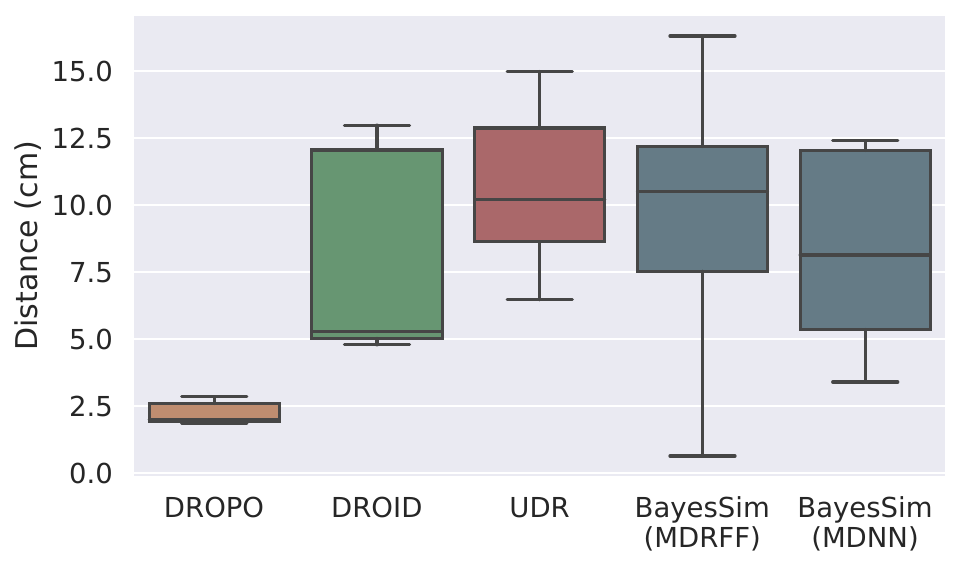}
    \caption{Sim-to-real performance in terms of the final distance to the target position on the pushing setup (in centimeters; lower is better).}
    \label{fig:panda_simtoreal}
\end{figure}

The converged DR distributions obtained from this data by all benchmarked methods are presented in Table~\ref{tab:sim2real_push_bounds}, averaged over three seeds.
Once again, we point out the smaller standard deviations DROID converged to, compared to DROPO. However, as the real-world dynamics parameters are not known, the final evaluation has to be carried out by means of policy performance on the real system, displayed in Figure~\ref{fig:panda_simtoreal} (5 rollouts per policy, per seed).

In this task, DROPO consistently outperformed previous methods and pushed the box as close as 2cm away from the target location, showing another successful zero-shot transfer. We observe how BayesSim, which resulted in less accurate parameter inference across our experiments, achieved inconsistent performance and obtains similar results to DROID and UDR on average. This result further motivated the need to infer dynamics such as to account for both accurate system identification and domain randomization at the same time, even for a simple pushing task.
BayesSim with neural network features (MDNN) predicted the box center of mass to be over 20 cm away from the center in each direction, with a standard deviation of over 50cm, which results in the center of mass being located way outside of the object.
Like in the Hopper experiment with an unmodeled phenomenon, this likely originates from the rather poor generalization to real-world data, which lies outside of the BayesSim network's training distribution---consisting of simulated data only.
However, since the environment implementation clips the center of mass values to be within object boundaries, the agent was still able to learn a somewhat sensible policy for some random seeds.
BayesSim with random Fourier features (MDRFF) shows slightly better generalization to real world data, producing a more sensible center of mass value, albeit with an excessive standard deviation of over 7cm in both directions and (as visualized in the detailed results in \ref{sec:bs:panda}) with negative mass being predicted on some occasions.
In terms of task performance, the trained policy still performs worse on average than DROPO or DROID.

\begin{table}[]
\scriptsize
\centering
\begin{tabular}{lllllll}
\hline
\multicolumn{1}{c}{}                                                    &     & $m$   & $f_x$  & $f_y$   & $com_x$  & $com_y$  \\ \hline
\multirow{2}{*}{\begin{tabular}[c]{@{}l@{}}Search\\ space\end{tabular}} & min & 0.08   & 0.20   & 0.20    & -0.032 & -0.032 \\
                                                                        & max & 2.00   & 2.00   & 2.00    & 0.032  & 0.032  \\ \hline
\multirow{2}{*}{\begin{tabular}[c]{@{}l@{}}Ground\\ truth\end{tabular}}                                            & $\mu^{*}$   & 0.7      & 0.8      & 0.8       & 0.01      & 0.01      \\
                                                                        & $\sigma^{*}$   & 0.3      & 0.2      & 0.2       & 0.001      & 0.001      \\ \hline
\multirow{2}{*}{\begin{tabular}[c]{@{}l@{}}DROPO\\ $^{\epsilon=0}$\end{tabular}} & $\mu^{*}$  & 0.788  & 0.603  & 0.683   & 0.015  & 0.009      \\
                                                                        & $\sigma^{*}$   & 0.108 & 0.033 & 0.18 & 0.001 & 0.003      \\ \hline
\multirow{2}{*}{DROID}                                                  & $\mu^{*}$   & 0.774      & 0.737      & 0.716       & 0.014      & 0.011      \\
                                                                        & $\sigma^{*}$   & 3.1e-11 & 1.3e-10 & 1.2e-11 & 7.6e-11 & 2.4e-11  \\ \hline
\multirow{2}{*}{BayesSim}                                               & $\mu^{*}$   & 0.703      & 0.894      & 0.840       & -0.011      & 0.004      \\
                                                                        & $\sigma^{*}$   & 0.269 & 0.224 & 0.257 & 0.309 & 0.234  \\ \hline
\end{tabular}
\caption{Results of the sim-to-sim pushing task experiment with randomized box mass $m$, table-box friction coefficients ($f_x,f_y$) and box center of mass in meters ($com_x,com_y$).}
\label{tab:sim2sim_push_bounds}
\end{table}

\begin{table}[]
\scriptsize
\centering
\begin{tabular}{llrrrrr}
\hline
\multicolumn{1}{c}{}                                                                  &              & \multicolumn{1}{c}{$m$} & \multicolumn{1}{c}{$f_x$} & \multicolumn{1}{c}{$f_y$} & \multicolumn{1}{c}{$com_x$} & \multicolumn{1}{c}{$com_y$} \\ \hline
\multirow{2}{*}{\begin{tabular}[c]{@{}l@{}}Search\\ space\end{tabular}}               & min          & 0.08                    & 0.20                      & 0.20                      & -0.032                      & -0.032                      \\
                                                                                      & max          & 2.00                    & 2.00                      & 2.00                      & 0.032                       & 0.032                       \\ \hline
\multirow{2}{*}{\begin{tabular}[c]{@{}l@{}}DROPO\\ $^{\epsilon = 1e-4}$\end{tabular}} & $\mu^{*}$    & 1.065                   & 0.387                     & 0.921                     & 0.012                       & -0.032                      \\
                                                                                      & $\sigma^{*}$ & 1.3e-04                 & 3.0e-03                   & 5.3e-04                   & 5.2e-05                     & 7.5e-04                     \\ \hline
\multirow{2}{*}{DROID}                                                                & $\mu^{*}$    & 0.826                   & 0.444                     & 0.665                     & -0.028                      & -0.028                      \\
                                                                                      & $\sigma^{*}$ & 1.4e-07                 & 2.3e-08                   & 2.6e-08                   & 7.7e-08                     & 6.6e-08                     \\ \hline
\multirow{2}{*}{\begin{tabular}[c]{@{}l@{}}BayesSim\\ MDNN model\end{tabular}}        & $\mu^{*}$    & 1.548                   & 3.889                     & 0.964                     & -0.322                      & -0.240                      \\
                                                                                      & $\sigma^{*}$ & 2.017                   & 0.805                     & 0.953                     & 0.577                       & 0.632                       \\ \hline
\multirow{2}{*}{\begin{tabular}[c]{@{}l@{}}BayesSim\\ MDRFF model\end{tabular}}       & $\mu^{*}$    & 0.796                   & 1.181                     & 1.013                     & -0.023                      & -0.040                      \\
                                                                                      & $\sigma^{*}$ & 0.252                   & 0.218                     & 0.316                     & 0.090                       & 0.075                       \\ \hline
\end{tabular}
\caption{Optimized dynamics distributions on a single offline trajectory collected on the real pushing environment. The converged box mass $m$, friction coefficients $f_{x}$ and $f_{y}$, and box center of mass $com_{x}$ and $com_{y}$ (in meters) are reported.}
\label{tab:sim2real_push_bounds}
\end{table}

\section{Conclusions}
\label{sec:conclusions}
In this paper we introduced DROPO, a method for crossing the reality gap by optimizing domain randomization distributions with limited, offline data.
We demonstrated that, unlike previous methods, DROPO is capable of accurately recovering the dynamics parameter distributions used to generate a dataset in simulation.
We also showed how DROPO may compensate for a misidentified value of a physical parameter, leading to well-performing reinforcement learning policies in the target domain.

We then moved on to real-world robotics setups---a hockey puck sliding and a box pushing tasks.
We demonstrated that a policy trained in simulation with DROPO can be directly transferred to the real world on both setups, while previous methods fail on the box pushing task.
We conclude that our method is able to effectively perform domain randomization and system identification at the same time, finding the right balance of parameter uncertainty by adjusting $\epsilon$.

In the current work, we used a variety of data collection strategies to collect the offline dataset---a partially-trained task policy (\textit{Hopper}), random sliding trajectories (\textit{Hockeypuck}) and human demonstrations obtained through kinesthetic teaching (\textit{Pushing}).
Future works may investigate how the data collection strategy impacts the accuracy of parameters obtained by offline methods such as DROPO and DROID;
in particular, prior work on exploration in meta-learning would make an interesting extension~\cite{arndt2021domaincuriosity,zhang2020learn}.

Under the current formulation, DROPO is limited to relatively low-dimensional parametric distributions.
This limitation stems from the rather costly non-differentiable optimization procedure; using less restrictive sampling procedures, such as Stein variational gradient descent used in~\cite{mehta20active}, could be a promising way of further improving the accuracy.
We showed that DROPO is capable of performing optimization in up to 18-dimensional parameter spaces; while this was enough for simple manipulation tasks, complex scenarios involving interactions between multiple objects may involve many more parameters, which could make DROPO prohibitively expensive.
Furthermore, in addition to larger dynamics parameter spaces, we encourage future works to extend our method for likelihood computation in high-dimensional state spaces, e.g. allowing DROPO to work on vision-based policies and plain images collected offline.
In this case, the general idea behind DROPO could be used not only to optimize the dynamics parameters of the system, but also to adjust visual domain randomization distributions for image rendering, for instance by adjusting parameters used for procedural generation of textures.

Domain randomization methods require the user not only to specify the dynamics parameter distributions, but also to select what parameters need to be randomized.
In our experiments, those parameters were indeed selected manually based on domain knowledge.
While this approach works well with simple and well-understood environments, it can present a challenge in situations where expert knowledge of the application domain is limited.
Even though randomizing as much as possible may appear to be a good idea on paper, it often leads to a significant increase in the computation cost. 
Additionally, some parameter combinations may be physically impossible, leading to simulator instabilities.

Aside from that, sim-to-real transfer with domain randomization often requires a significant development effort, requiring an accurate physics engine, object meshes and parametric simulation models to be available.
As of now, there is no automated way of designing simulation models that accurately mimic real-world setups; such methods would make the initial software development effort much lower, allowing sim-to-real transfer with domain randomization to be used more widely in the industry, for example in digital twin applications.
Recent developments in automated design open up the possibility of automating the construction of simulation models beyond parametric inference. 
More specifically, the object shapes and the general kinematic structure of the scene could be inferred based on the appearance of the real world system, and have its dynamics parameters tuned based on pre-collected physical observations using a method like DROPO.

\section*{Acknowledgment}
\label{sec:acknowledgment}
This work was supported by Academy of Finland grants 317020 and 328399. We acknowledge the computational resources generously provided by CSC – IT Center for Science, Finland, and by the Aalto Science-IT project.



\bibliographystyle{ieeetr}
\bibliography{main_copied.bib}


\appendix
\section{Least squares for offline DR optimization}
\label{sec:appendix:least_squares_adr}
In this appendix section we attempt to give a more thorough explanation to why DROID converged to point-estimate dynamics parameters across all our experiments.
In particular, we aim to provide an intuition for why we discourage the use of least squares-like methods for trajectory alignment during \textit{offline} optimization of DR distributions. This discussion is further motivated by the results reported in~\cite{calibratingmehta}, where SimOpt's~\cite{chebotar19closing} objective function was used in the offline setting by replaying actions from off-policy data in simulation.
Similarly to our results with DROID, this approach resulted in point-estimate dynamics estimates, leading the authors to consider offline SimOpt as a point-estimate system identification method.

Based on the notation used throughout our work, let us consider the optimization problem aiming to minimize the following objective function:

\begin{equation} \label{appeq:objfun}
    J( \xi ) =\frac{1}{K}\sum _{i=1}^{K}\Vert f( s_{i} ,a_{i} ;\xi ) -s_{i+1}\Vert ^{2} 
\end{equation}
 
with $f(\cdot; \xi)$ being the simulator dynamics---assumed to be deterministic for simplicity---parameterized by parameters $\xi$.

Note that, since the objective function is defined for a given dynamics parameter vector $\xi$, minimizing it by acting on the full distribution is not straightforward. For example, DROID~\cite{tsai2021droid} uses CMA to minimize the squared residuals and then retains the final inner CMA covariance matrix to ultimately build the DR distribution.
Similarly, SimOpt~\cite{chebotar19closing} uses a sampling-based gradient-free algorithm based on relative entropy policy search~\cite{Peters2010RelativeEP} allowing to minimize residuals using only samples $\xi \sim p_\phi(\xi)$.

Regardless of the specific method, since the final goal is the minimization of $J(\xi)$, we can reason about its expected value w.r.t. $p_\phi(\xi)$, which we are ultimately interested in minimizing. Intuitively, when optimizing the objective in~(\ref{appeq:objfun}) under no additional assumptions or constraints, it is indeed reasonable to believe that no benefit is gained by ending up with a higher expected value of $J(\xi)$ in exchange for a certain desired variance.

We then claim that optimization problems aiming to optimize the expected value of quantity in~\ref{appeq:objfun} by acting on the random component itself, inevitably converge to point-estimates. Interestingly enough, this draws similarities to the weighted average of a finite sequence of numbers: no matter how you pick the weights (e.g. the probability distribution), the average will always be greater than or equal to the minimum of the sequence; only when moving all the probability mass to the minimum value itself we get the lowest possible average value.

We attempt to give a concise mathematical demonstration for this in our continuous case, under a few assumptions for the sake of simplicity.
Let us bound the search space of the optimization problem to an arbitrary region in the dynamics space. Let $j_{min}=J(\xi_{min})$ be the global minimum of $J(\xi)$ in the region delimited by the support of the current $p_\phi(\xi)$, potentially trimmed by the predefined search space in case of unbounded support.
Therefore, note how the random variable $Y = J(\xi) - j_{min}$ is always greater than or equal than zero. Following the non-negativity and linearity properties of expected values, we consequently get:

\begin{equation} \label{appeq:nonnegativity}
    \mathbb{E}[ Y] \ =\mathbb{E}[ J( \xi ) -j_{min}] \ \geqslant \ 0\\
\end{equation}

\begin{equation} \label{appeq:linearity}
\mathbb{E}[ J( \xi )] \geqslant j_{min}
\end{equation}

This final result indicates that regardless of how we design the distribution $p_\phi(\xi)$, we cannot do better than simply minimizing the objective function w.r.t. $\xi$ only --- indeed, we can only do worse. Therefore, the optimization problem should converge to the point-estimate $\xi_{min}$ even if acting on the full distribution, assuming that our optimizer converges to the global minimum.

We then encourage future works to take these considerations into account when designing optimization of dynamics distributions with fixed offline data. For example, one could add custom constraints to promote variance in the dynamics parameters or, as shown by our work and BayesSim~\cite{bayessimramos}, move to probabilistic metrics to infer target parameters.

Finally, it's worth noting that these considerations do not apply to the original SimOpt method (as presented in~\cite{chebotar19closing}), as its objective function is conceptually different in the online setting: SimOpt optimizes $\phi$ so that \emph{a policy} trained on such distribution gets minimum squared residuals between its observed sim and real states; i.e., the objective is directly to get a policy that performs the same in both domains, as opposed to simple trajectory alignment.
Thus, having policy training and data collection from the real-world system in the loop prevents the optimization from converging to a point estimate, as a policy trained on a point estimate would---in most cases---fail to generalize to the real world, resulting in a larger discrepancy between simulated and real trajectories than a policy trained on a non-degenerate dynamics distribution.

\section{The impact of $\epsilon$}
\label{sec:appendix:epsilon}

\begin{figure}
    \centering
    \begin{subfigure}{0.8\linewidth}
    \centering
    \includegraphics[width=\linewidth]{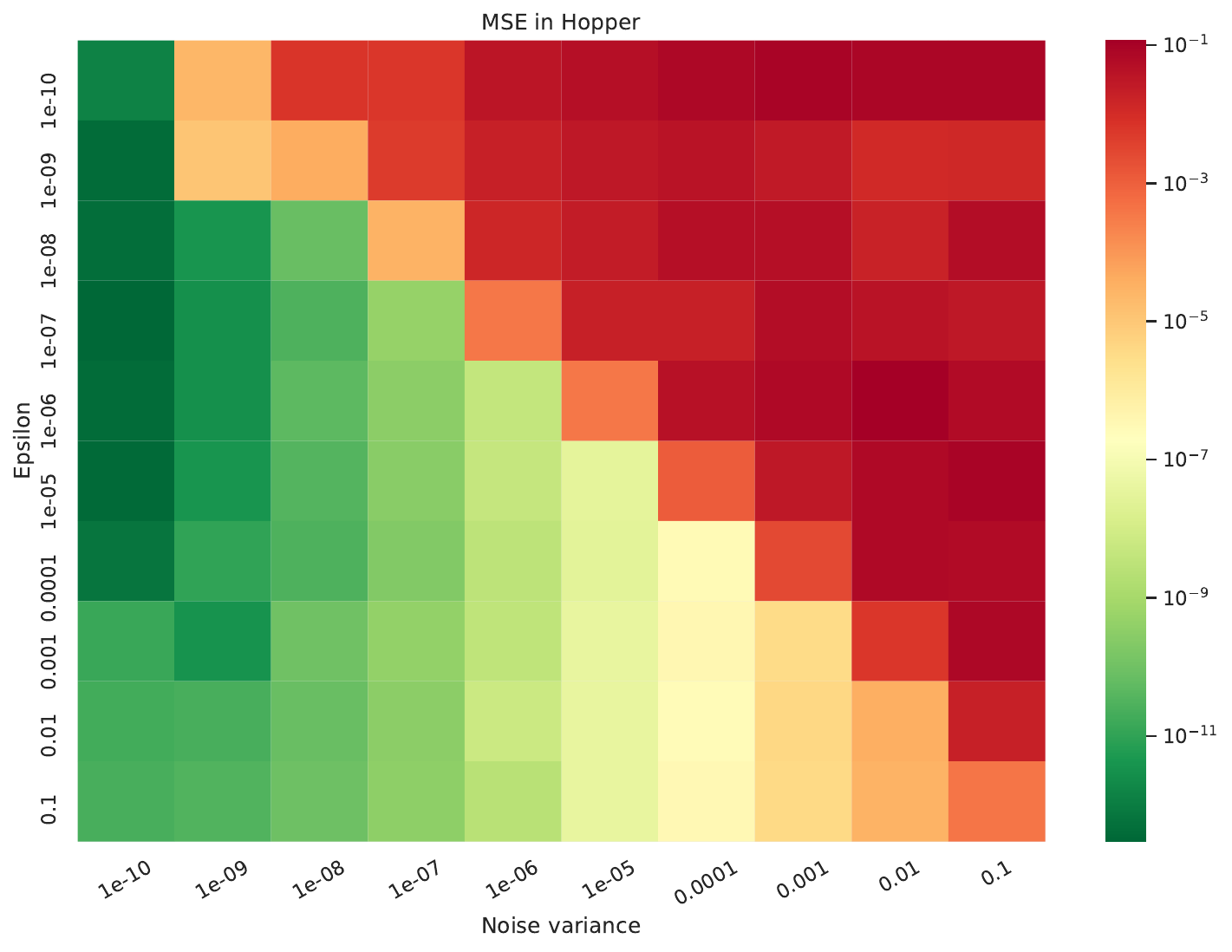}
    \caption{}
    \label{fig:hop_pointest_eps_noise_mse}
    \end{subfigure}
    
    \begin{subfigure}{0.8\linewidth}
    \centering
    \includegraphics[width=\linewidth]{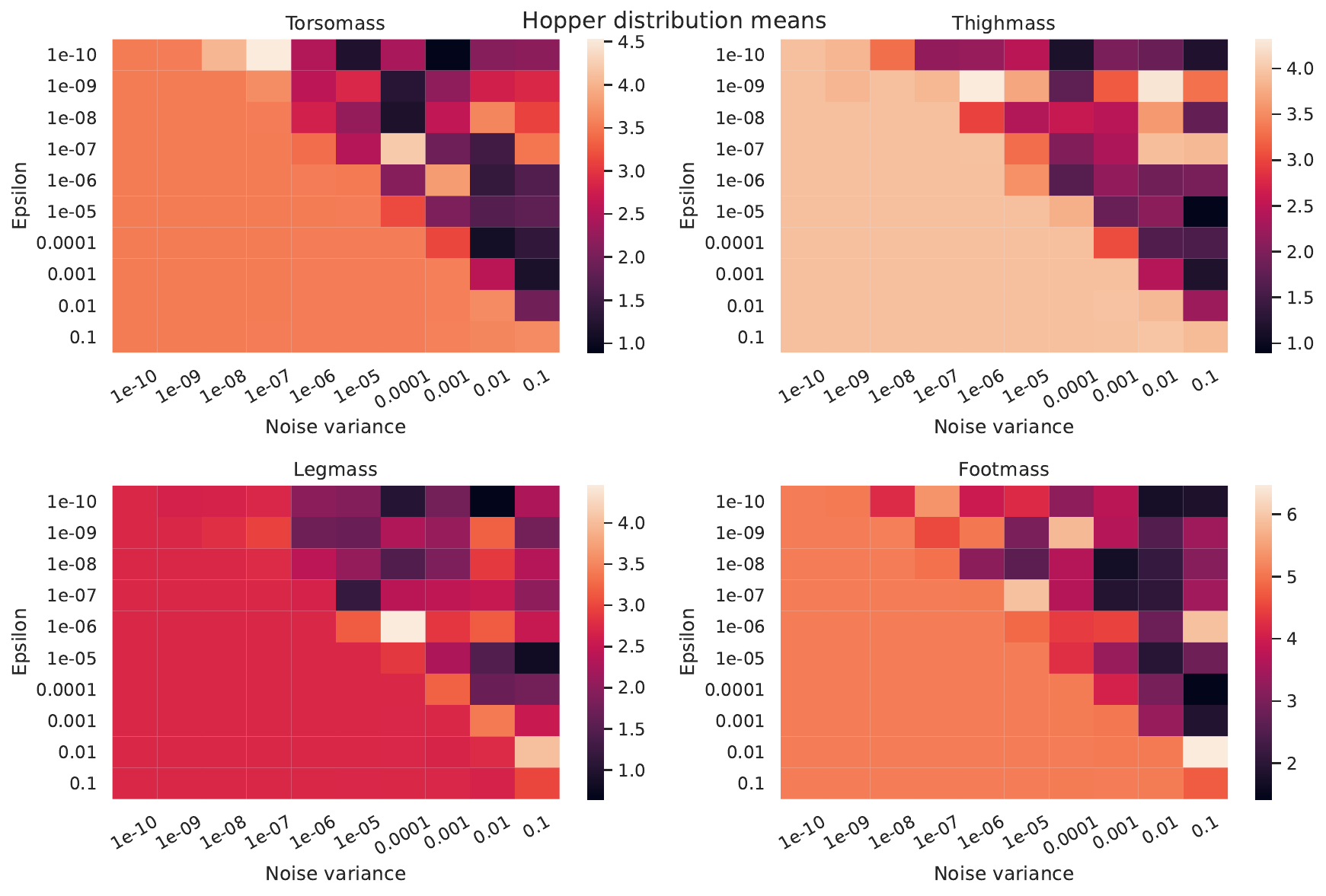}
    \caption{}
    \label{fig:hop_pointest_eps_noise_means}
    \end{subfigure}
    \begin{subfigure}{0.8\linewidth}
    \centering
    \includegraphics[width=\linewidth]{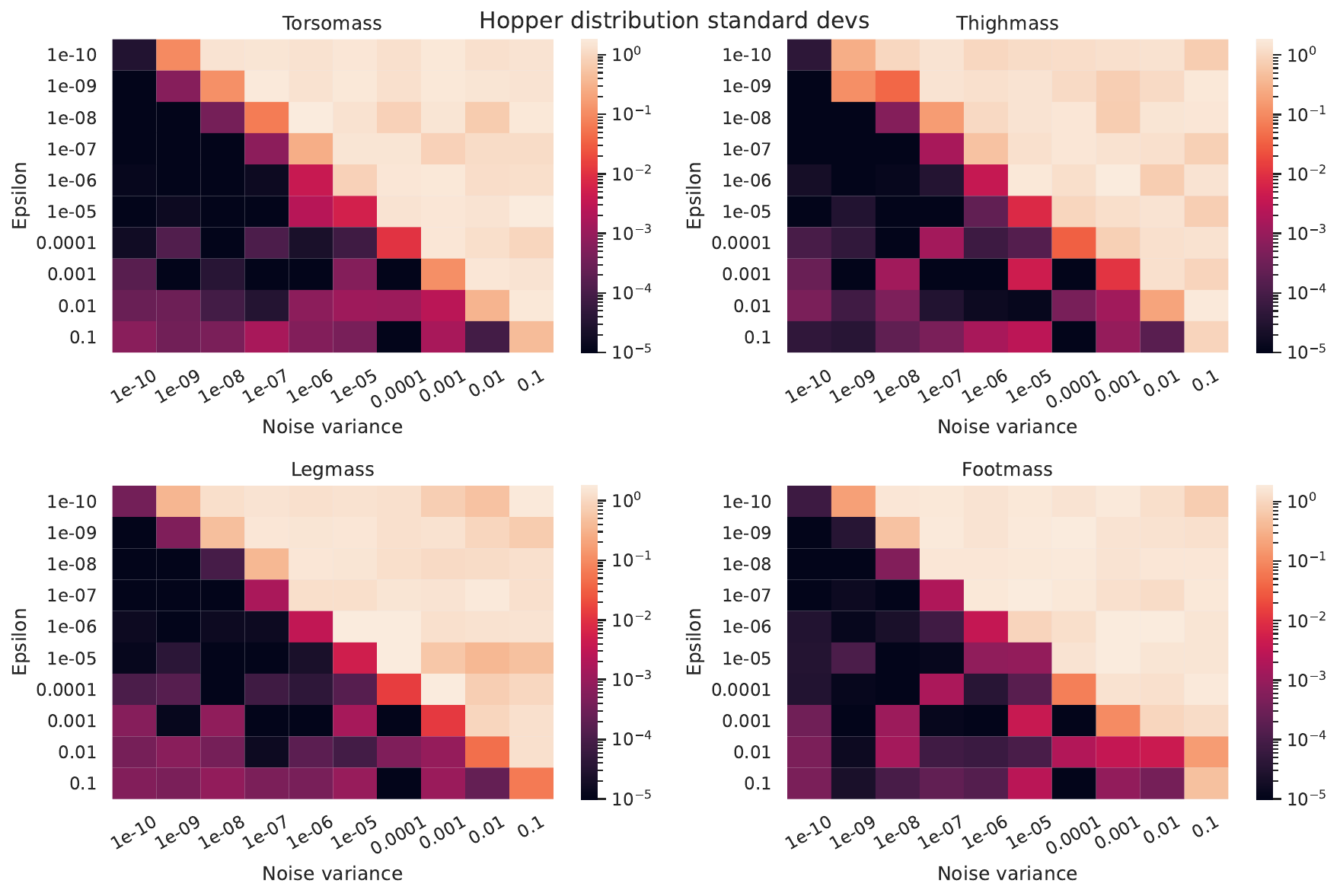}
    \caption{}
    \label{fig:hop_pointest_eps_noise_stdevs}
    \end{subfigure}
    \caption{Relationship between noise level, epsilon and the resulting (\subref{fig:hop_pointest_eps_noise_mse}) MSE and (\subref{fig:hop_pointest_eps_noise_means},~\subref{fig:hop_pointest_eps_noise_stdevs}) distribution, expressed by mean and standard deviation in the Hopper point-dynamics system identification case. (results averaged over 4 random seeds)}
    \label{fig:epsilon_noise_sweep_pointest}
\end{figure}

To further analyze the impact of $\epsilon$ on DROPO's performance in presence of noise, we performed a secondary analysis where we corrupt the collected Hopper datasets with different levels of noise.

The results for the point-dynamics Hopper dataset described in Section~\ref{sec:point_est_hopper} are depicted in Figure~\ref{fig:epsilon_noise_sweep_pointest}.
In line with the main experimental evaluation, we observed a consistent drop in MSE where $\epsilon$ matches the noise level $\sigma^2$. In addition, the results in terms of the converged mean (Figure~\ref{fig:hop_pointest_eps_noise_means}) are consistent for $\epsilon \geq \sigma^2$, and the standard deviations are very low, corresponding to a correct identification of point dynamics parameters.

We finally performed a similar analysis for the distribution recovery case, which produced similar results: the dynamics distribution recovery is most accurate at $\epsilon$ close to the noise variance $\sigma^2$, which also corresponds to the sharp drop in MSE. In particular, note how the converged standard deviations of the four masses resemble the ground truth distributions defined in Section~\ref{sec:distribution_recovery}, as long as $\epsilon$ does not increase to overly large values w.r.t. the current noise level.

\begin{table*}[]
    \scriptsize
    \centering
    \begin{tabular}{c|c|c|c||c|c}
         Experiment & \multicolumn{1}{p{2cm}|}{\centering Number of\\trajectories} & \multicolumn{1}{p{2cm}|}{\centering Transitions\\per trajectory} & \multicolumn{1}{p{2cm}||}{\centering Time per\\transition} &  \multicolumn{1}{p{2cm}|}{\centering Total\\transitions} & \multicolumn{1}{p{2cm}}{\centering Total\\time} \\
         \hline
         Hopper & 2 & 500 & 8ms & 1000 & 8s \\
         Hockeypuck & 5 & 750 & 10ms & 3750 & 37.5s \\
         Sim-to-sim push & 50 & 30 & 20ms & 1500 & 30s \\
         Sim-to-real push & 1 & 1286 & 10ms & 1286 & 12.86s
    \end{tabular}
    \caption{The amount of data used for each experiment, broken down into trajectories, state transitions, and the total time.}
    \label{tab:amount_of_data}
\end{table*}

\begin{figure}
    \centering
    \begin{subfigure}{0.8\linewidth}
    \centering
    \includegraphics[width=\linewidth]{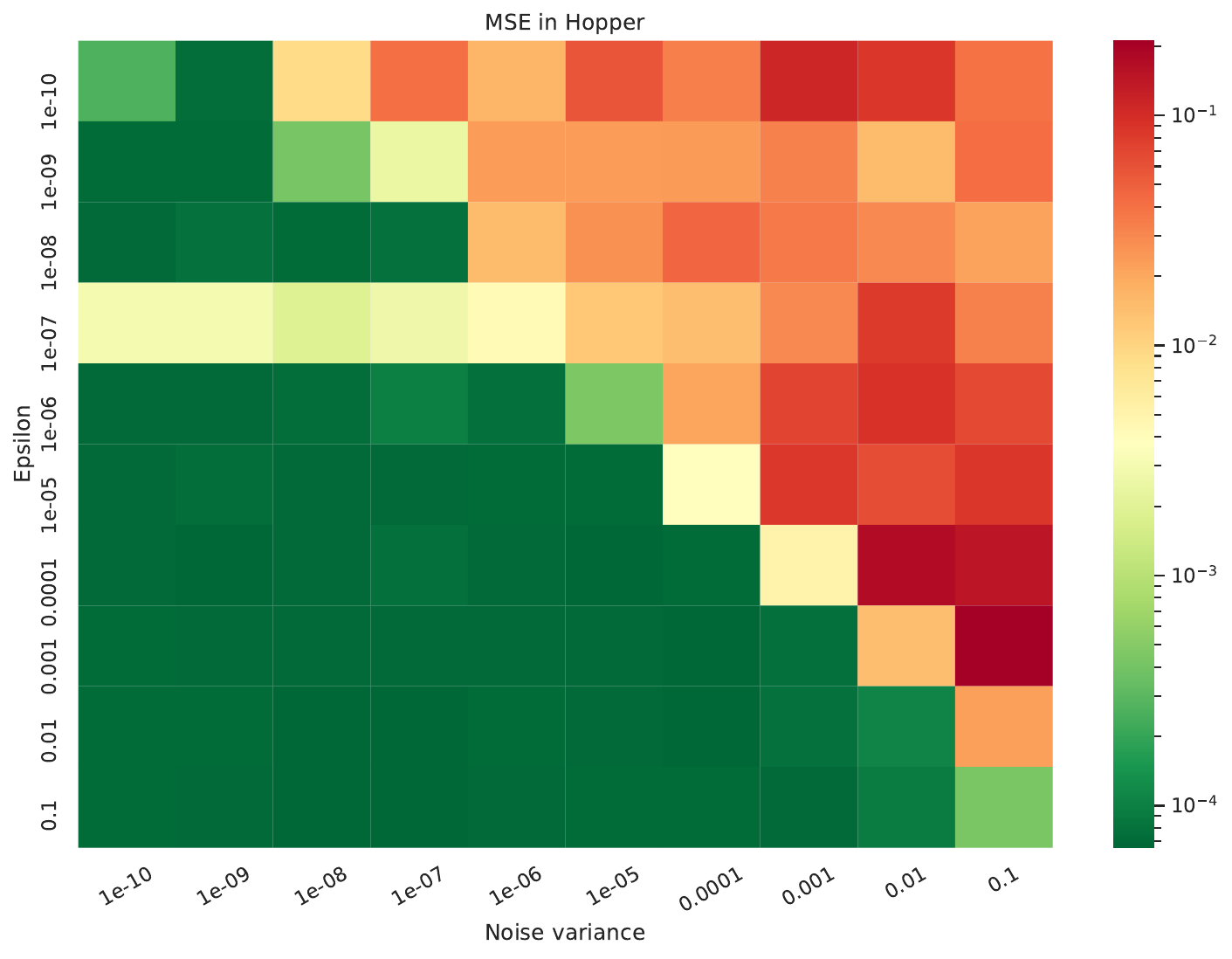}
    \caption{}
    \label{fig:hop_distrec_eps_noise_mse}
    \end{subfigure}
    \begin{subfigure}{0.8\linewidth}
    \centering
    \includegraphics[width=\linewidth]{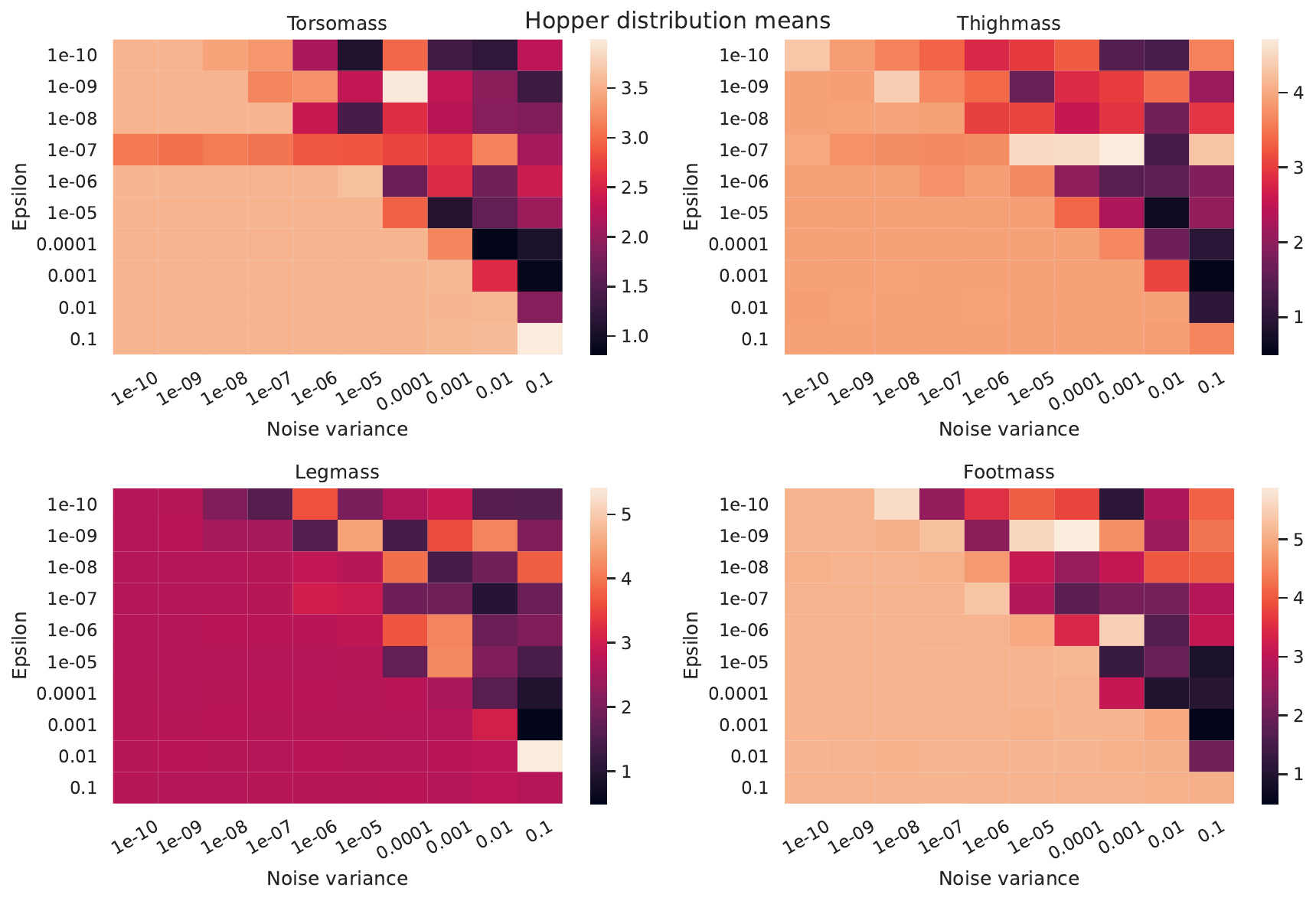}
    \caption{}
    \label{fig:hop_distrec_eps_noise_means}
    \end{subfigure}
    \begin{subfigure}{0.8\linewidth}
    \centering
    \includegraphics[width=\linewidth]{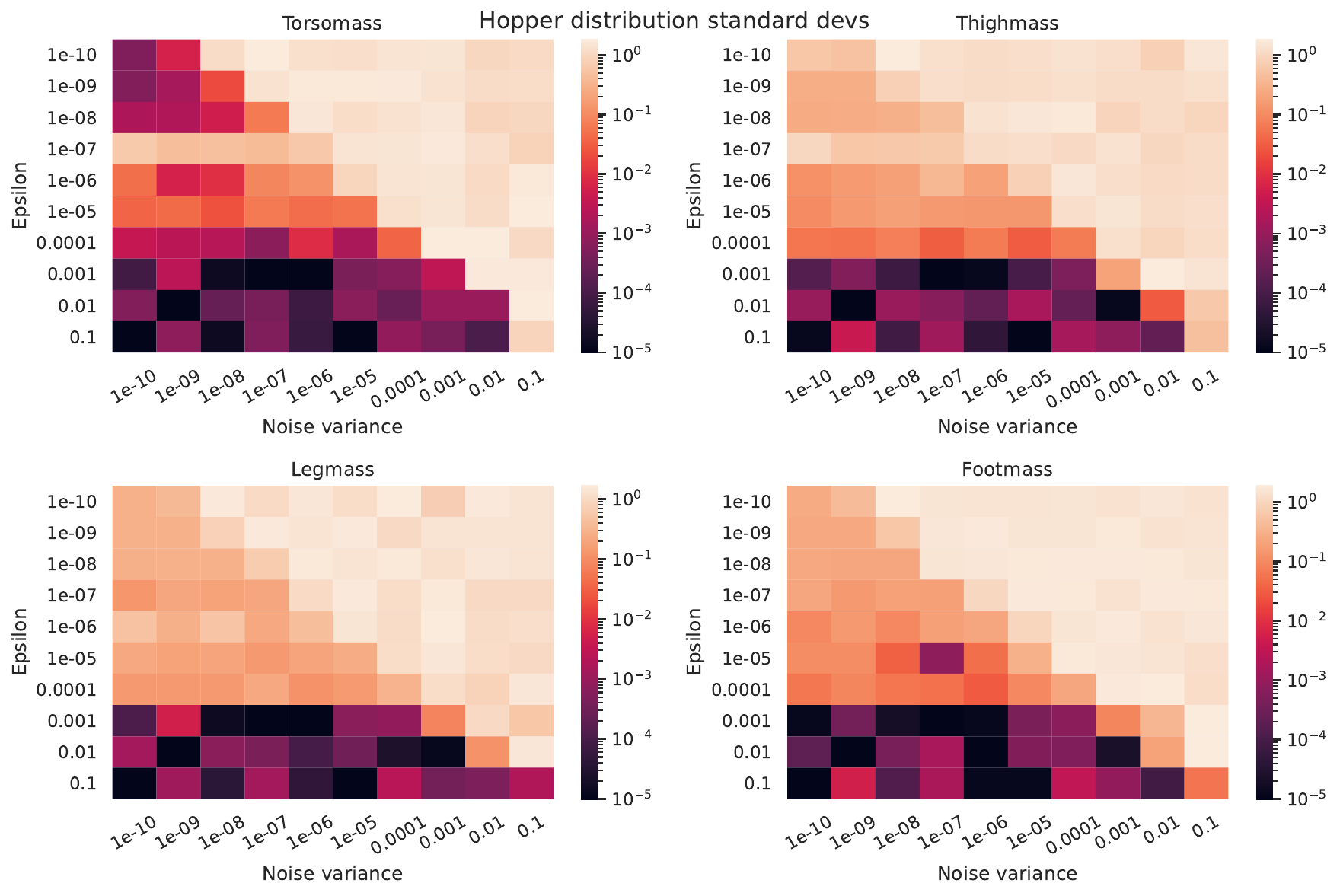}
    \caption{}
    \label{fig:hop_distrec_eps_noise_stdevs}
    \end{subfigure}
    \caption{Relationship between noise level, epsilon and the resulting (\subref{fig:hop_distrec_eps_noise_mse}) MSE and the (\subref{fig:hop_distrec_eps_noise_means},~\subref{fig:hop_distrec_eps_noise_stdevs}) distribution, expressed by mean and standard deviation in the Hopper dynamics distribution recovery case. (results averaged over 4 random seeds)}
    \label{fig:epsilon_noise_sweep_distrrec}
\end{figure}

\section{Sensitivity to the amount of data}
\label{sec:appendix:amountofdata}

\begin{figure}
    \centering
    \includegraphics[width=\linewidth]{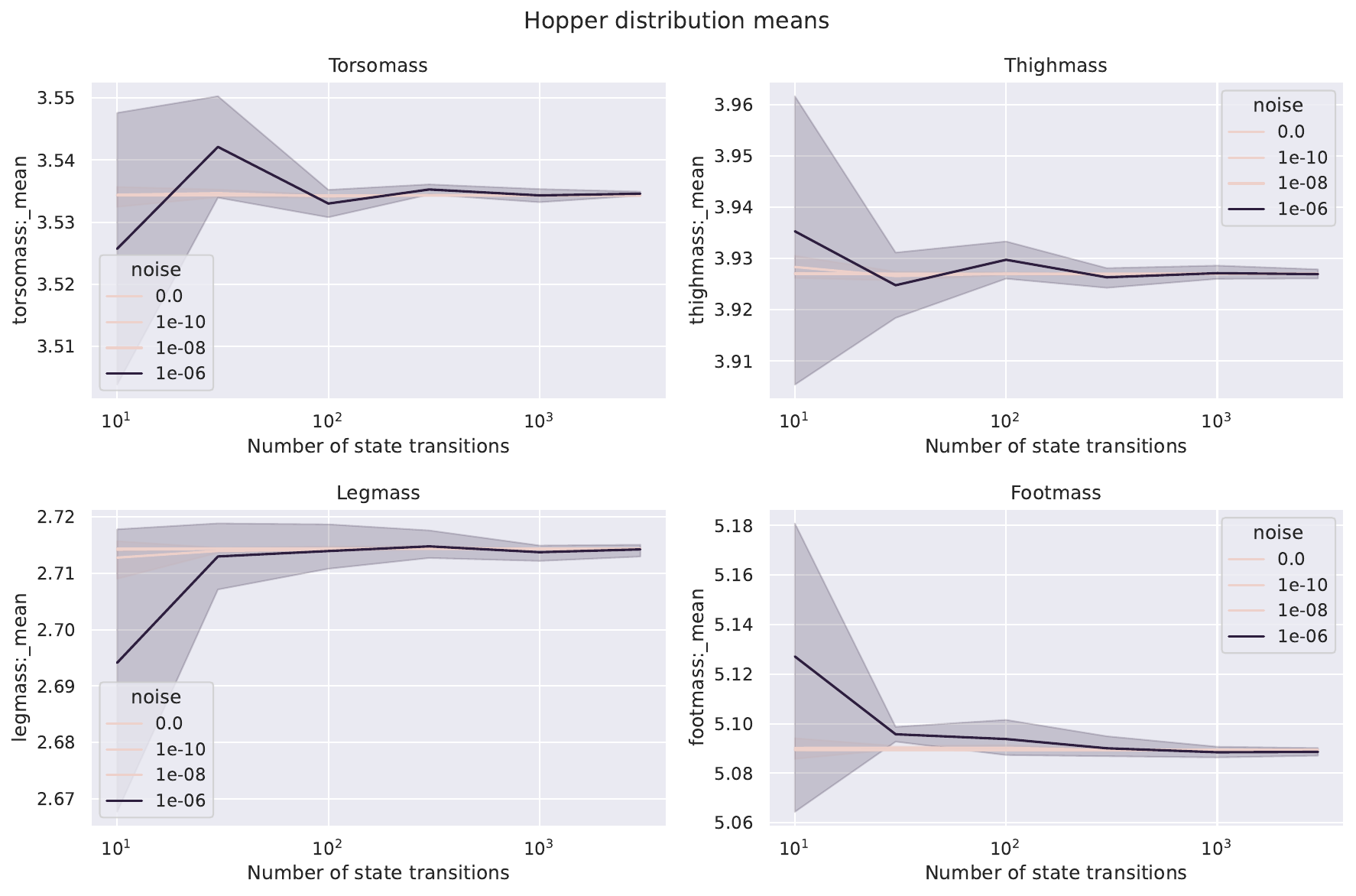}
    \caption{Hopper dynamics parameter estimated mean with its 95\% confidence interval (estimated by bootstrapping over 4 random seeds)}
    \label{fig:hop_means}
\end{figure}

As observed in our experiments section, DROPO succeeded to transfer RL pushing policies to the real world with as little as a single demonstrative trajectory for parameter inference. In this section, we aim to quantify the sensitivity of DROPO to the amount of offline data collected. To this end, we analyzed the converged means of the dynamics distributions with different amount of state transitions in different noise levels, for the Hopper environment in the point-dynamics setting of Section~\ref{sec:point_est_hopper}. In particular, we tested DROPO with 10, 30, 100, 300, 1000 and 3000 state transitions sub-sampled from the offline dataset, and kept $\epsilon$ fixed at the value of $10^{-5}$. We report the confidence interval of the converged mean of each mass of the Hopper in Fig.~\ref{fig:hop_means}, obtained from 4 runs for each evaluation.

We observe that, when noise variance is around $10^{-8}$ or lower, as little as 10 state transitions are enough for DROPO to converge to values very close to the ground truth values.
For higher noise levels, this number increases to several hundreds of state transitions. However, note the relatively small scale on the y-axis, where confidence intervals get as wide as 10 grams at most. 

For the sake of clarity, we finally combine and report in Table~\ref{tab:amount_of_data} the amount of data used in all our experiments.
Target data has been collected in less than a minute of interaction for all tasks, motivating the adoption of offline DR optimization for critical-to-operate environments. Noticeably more data has been used to infer dynamics for the Hockeypuck environment, due to the nature of the task and the robot arm moving through the air before hitting the hockey puck, with the puck moving only for about a second in each trajectory (thus, only a small subset of state transitions could be used to infer the dynamics of the hockey puck).

\section{Complete BayesSim results}
BayesSim parameterizes the resulting dynamics distributions as a mixture of multivariate Gaussians, while other methods used simple Gaussian distributions.
In order to clearly present the inferred distributions in comparison with the other methods in the Experiments section of the paper, we simplified the BayesSim results by only reporting the diagonal uncorrelated approximation of the Gaussian mixture.
While this gives a good intuition about the accuracy of the results, it loses some of the information; hence, we here provide plots of the full mixture Gaussian distributions.

In this section, we provide the full BayesSim inference results by plotting histograms and joint distributions over all pairs of dynamics parameters.
Each color corresponds to an independent training run of the BayesSim inference network.

\subsection{Hopper}
Here, we present the complete dynamics estimation results in Hopper. The ground-truth values for the point dynamics recovery experiment were reported in Table~\ref{tab:pointest_convergence} and for the dynamics distribution recovery in Figure~\ref{fig:distr_recovery_results}.

\begin{figure}[H]
    \centering
    \includegraphics[width=\linewidth]{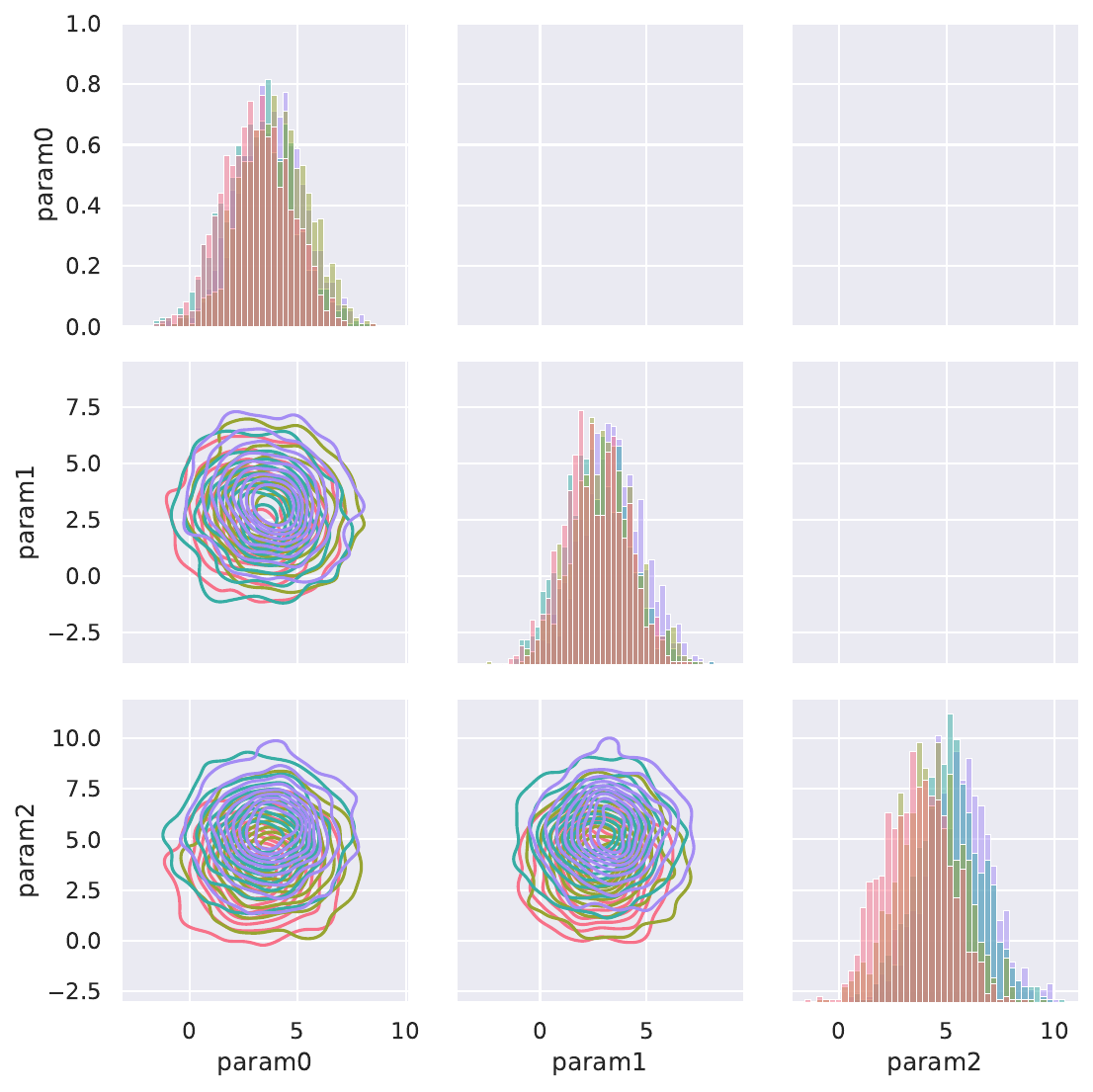}
    \caption{Complete dynamics estimation results in Hopper with unmodeled phenenomena with BayesSim}
    \label{fig:bs:hopper_wm}
\end{figure}

The parameters (from param0 to param3) represent the masses of consecutive Hopper links---torso, thigh, leg and foot.
In the unmodeled phenomenon experiment (Figure~\ref{fig:bs:hopper_wm}), the torso mass is excluded as it is fixed to an incorrect value and is not inferred during optimization.
In this case, as previously argued, the test environment does not lie within the distribution of environments used for training data generation. 
As a result, the evaluation data lies outside of the training domain of the BayesSim inference network.
This is likely the reason why the network produces a distribution with a large standard deviation centered around the mean of the search bounds.

We observe that in all distribution recovery experiments, BayesSim correctly returns a mixture model that is effectively a unimodal Gaussian distribution (with the weights of the remaining mixture components very close to zero).

\begin{figure}[H]
    \centering
    \includegraphics[width=\linewidth]{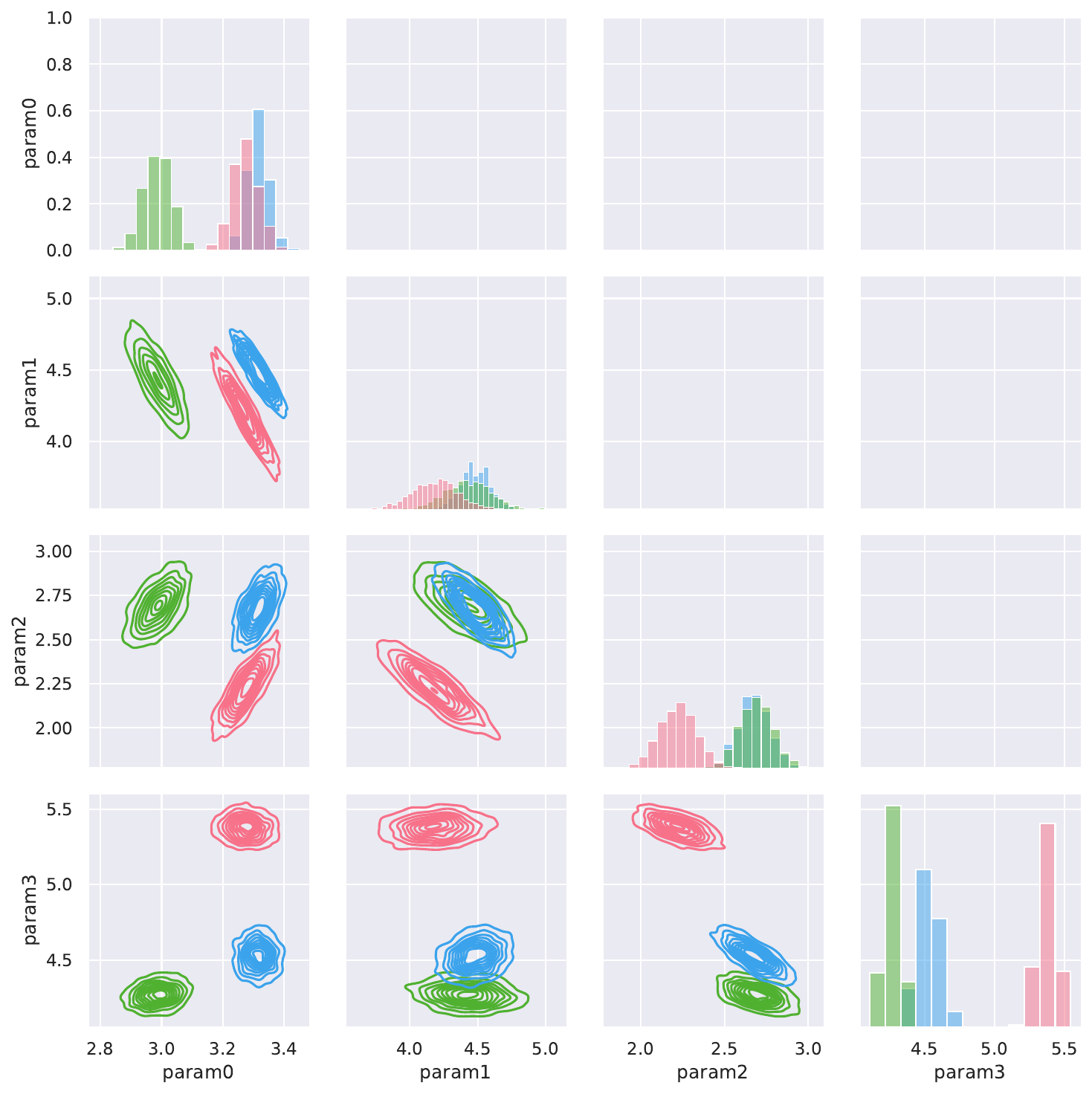}
    \caption{Complete dynamics distribution recovery results in Hopper with BayesSim}
    \label{fig:bs:hopper_dr}
\end{figure}

In Figure~\ref{fig:bs:hopper_dr}, we also observed a significant discrepancy between different repetitions of the experiment, i.e. independent training runs of the BayesSim inference network.

\begin{figure}[H]
    \centering
    \includegraphics[width=\linewidth]{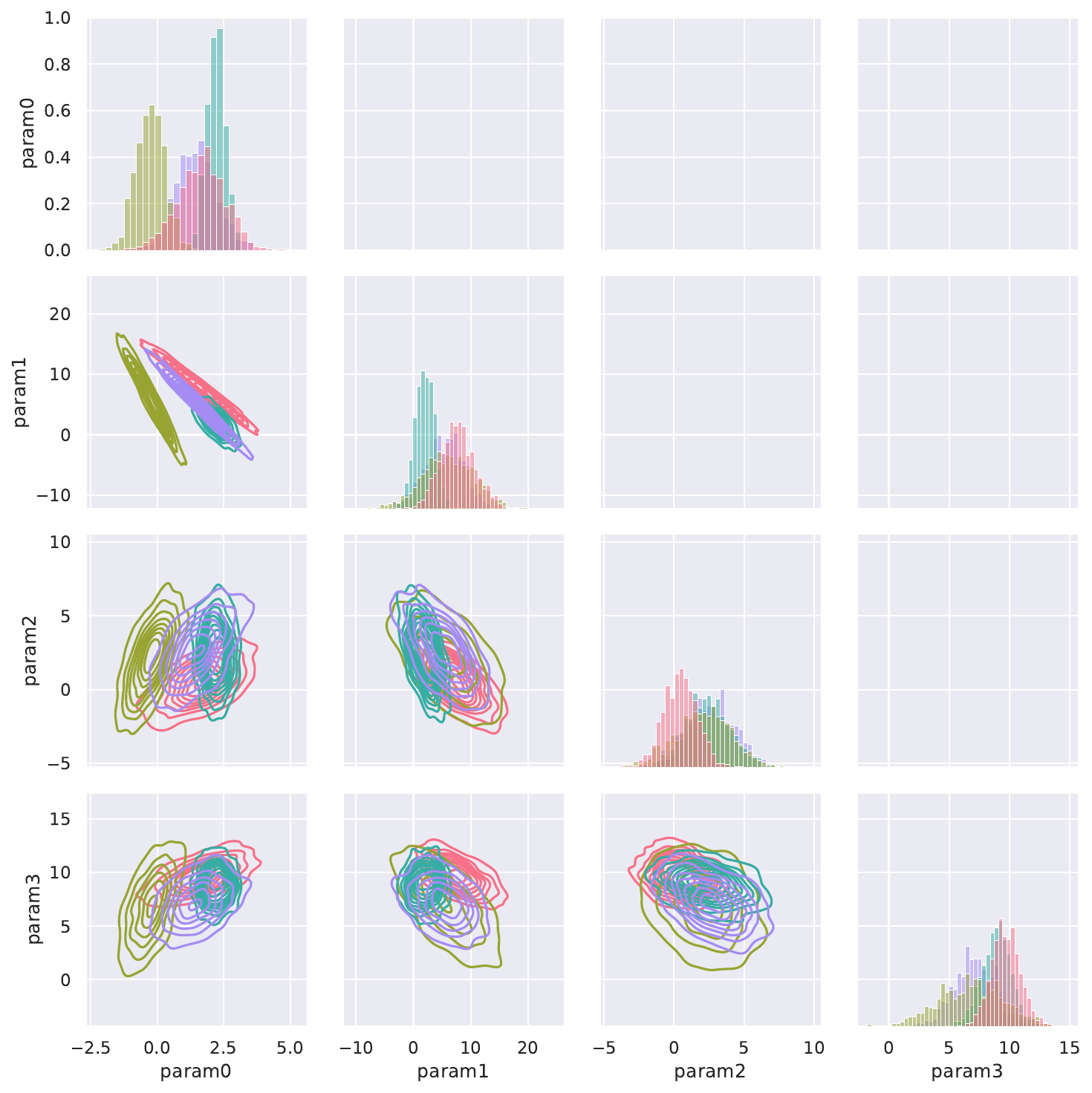}
    \caption{Complete dynamics identification results in Hopper with BayesSim}
    \label{fig:bs:hopper_pe}
\end{figure}

\subsection{Panda} \label{sec:bs:panda}

Here, we present the full BayesSim results for the Panda pushing task. The estimated dynamics parameters are, in order, the mass of the box, the friction coefficient values along axes $x$ and $y$, and the $x$ and $y$ position of center of mass of the box.

\begin{figure}
    \centering
    \includegraphics[width=\linewidth]{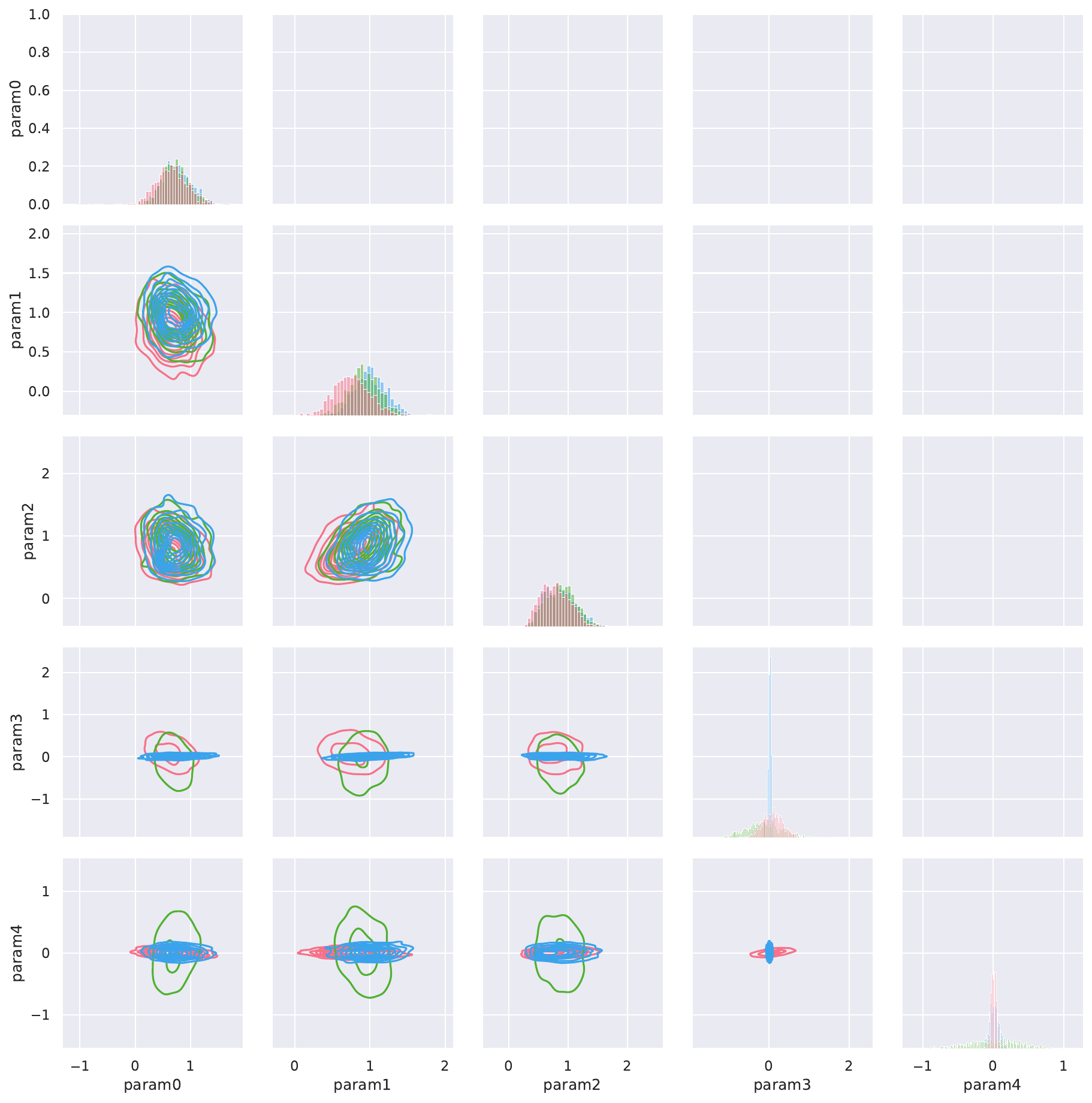}
    \caption{Complete dynamics distribution recovery results in Panda with BayesSim}
    \label{fig:bs:panda_dr}
\end{figure}

In the sim-to-sim inference experiment (Figure~\ref{fig:bs:panda_dr}) we observe that, despite there being some variance between random seeds, the models generally agree that the center of mass along $x$ and $y$ should be shifted in the positive direction, and that the mass of the object is rather hard to identify (as expressed by the large standard deviation for param0).

\begin{figure}
    \centering
    \includegraphics[width=\linewidth]{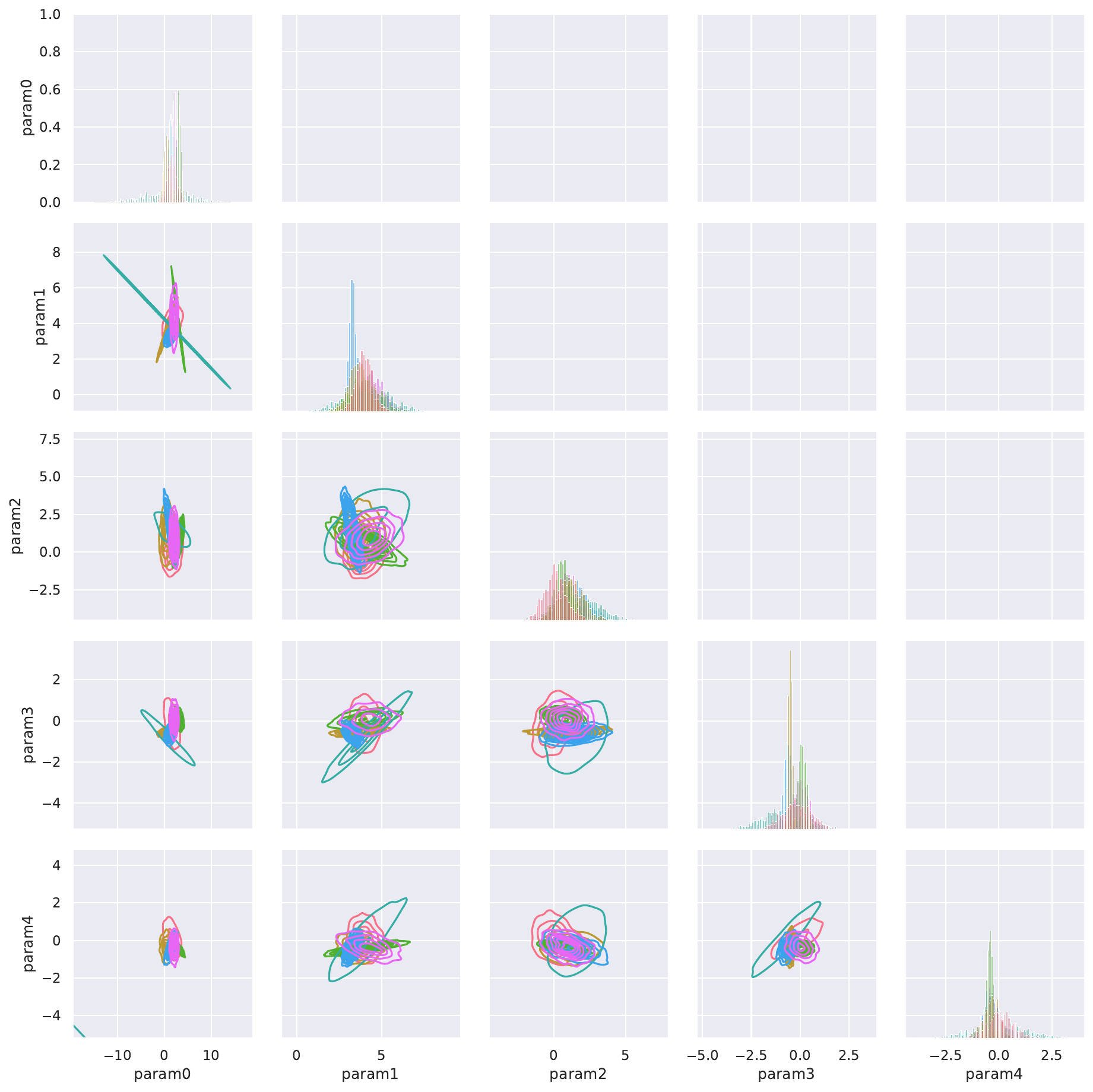}
    \caption{Complete real-world dynamics estimation results in Panda with BayesSim (MDNN features)}
    \label{fig:bs:real_panda_mdn}
\end{figure}

\begin{figure}
    \centering
    \includegraphics[width=\linewidth]{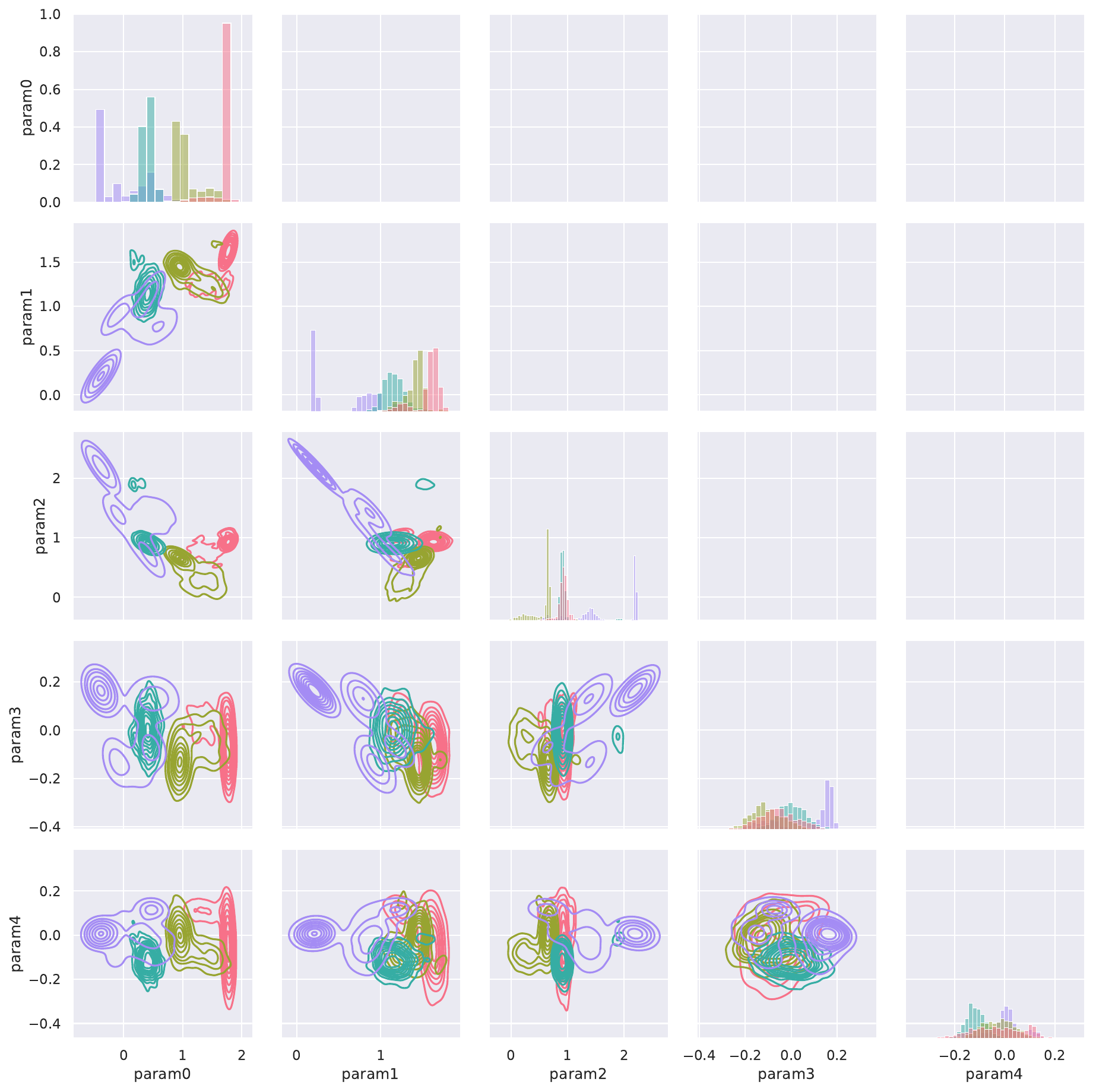}
    \caption{Complete real-world dynamics estimation results in Panda with BayesSim (MDRFF features)}
    \label{fig:bs:real_panda_mdrff}
\end{figure}

However, when operating on real-world data, the model's performance degrades significantly, especially with the neural network features (MDNN), shown in Figure~\ref{fig:bs:real_panda_mdn}.
In this case, the model produces very large standard deviations, potentially hindering the training process and often leading to unfeasible physical parameters---e.g. negative masses or box center of mass that lies outside the box boundaries.
To address this issue, all values were clipped inside the simulator and kept within the pre-identified search space, which made the policy training possible.
Like in the unmodeled Hopper experiment, we suspect that this behaviour is the result of the network being evaluated on out-of-distribution data. 
This hypothesis appears stronger when looking at the results obtained with the quasi-random Fourier features, where the inferred distributions look more sensible, although in some cases the mixture components still predict the mass to be negative (Figure~\ref{fig:bs:real_panda_mdrff}).
The feature extraction process reduces the dimensionality of the input data, making it easier for the model to generalize to new data, as it is more likely that the noisy, unmodeled real-world trajectory would appear close to data seen during training.

\end{document}